\documentclass{bioont}

\usepackage{natbib}
\usepackage{url}
\usepackage{xspace}
\usepackage{graphicx} 
\usepackage{longtable}

\begin{document}

\title[Can inferred provenance and its visualisation be used to detect
  erroneous annotation? A case study using UniProtKB.]{Can inferred
  provenance and its visualisation be used to detect erroneous
  annotation? A case study using UniProtKB.}

\author[M. J. Bell \textit{et~al}]{Michael J. Bell$^{1}$, Matthew
  Collison$^{2}$, Phillip Lord$^{1,}$\footnote{to whom correspondence
    should be addressed}}

\address{$^{1}$School of Computing Science, Newcastle University,
  Newcastle-Upon-Tyne, NE1 7RU, UK.\\ $^{2}$School of Chemical
  Engineering and Advanced Materials, Newcastle University,
  Newcastle-Upon-Tyne, NE1 7RU, UK.}

\date{August 2013}

\maketitle

\section*{Abstract}

A constant influx of new data poses a challenge in keeping the annotation in
biological databases current. Most biological databases contain significant
quantities of textual annotation, which often contains the richest source of
knowledge. Many databases reuse existing knowledge, during the curation
process annotations are often \emph{propagated} between entries. However, this
is often not made explicit. Therefore, it can be hard, potentially impossible,
for a reader to identify where an annotation originated from.

Within this work we attempt to identify annotation \emph{provenance}
and track its subsequent propagation. Specifically, we exploit
annotation reuse within the UniProt Knowledgebase (UniProtKB), at the
level of individual sentences. We describe a visualisation approach
for the provenance and propagation of sentences in UniProtKB which
enables a large-scale statistical analysis. Initially levels of
sentence reuse within UniProtKB were analysed, showing that reuse is
heavily prevalent, which enables the tracking of provenance and
propagation. By analysing sentences throughout UniProtKB, a number of
interesting propagation patterns were identified, covering over
$100,000$ sentences. Over $8000$ sentences remain in the database
after they have been removed from the entries where they originally
occurred. Analysing a subset of these sentences suggest that
approximately $30\%$ are erroneous, whilst $35\%$ appear to be
inconsistent. These results suggest that being able to visualise
sentence propagation and provenance can aid in the determination of
the accuracy and quality of textual annotation.

Source code and supplementary data are available from the authors
website\footnote{http://homepages.cs.ncl.ac.uk/m.j.bell1/sentence\_analysis/}.

\section*{Introduction}

Biological databases store, organise and share ever-increasing
quantities of data~\cite{Robbins94Biological}. In addition to storing
raw biological data, such as protein sequences, many databases aim to
attach \emph{textual annotation} to a given database entry. This
textual annotation provides a mechanism to convey understanding of
the underlying biology, providing information such as protein function
and subcellular location. In describing the current knowledge about
the database entry, textual annotations can form the foundations for
further research~\cite{Experimental07Buza} emphasising their crucial
role in biological databases.

The quality and correctness of textual annotations inevitably varies
between databases and entries. This can depend on many factors, such
as: the current evidence supporting the function of the protein; the
curation and review process; and the curators' judgement in extracting
information from biomedical literature~\cite{Hong08Gene,
  Dolan05Procedure}. The kind of metadata describing annotations also
varies between databases and entries, limiting the ability to compare
them. For example, the source (or \emph{provenance}) and last updated
date of a Gene Ontology (GO) annotation is not always
apparent~\cite{Buza08Gene}.

At the highest level, we can distinguish between two types of
annotation curation process: manual curation and automated curation.
It is generally held that manual curation is of higher quality and
correctness than its automated counterpart. This is mainly because
expert curators have the ability to access, evaluate and interpret a
wide range of scientific literature as a source of information for
annotations (as is the case for
UniProtKB/Swiss-Prot~\cite{Magrane11UniProt} and
FlyBase~\cite{Drysdale05FlyBase}). However, automated annotation
pipelines, such as UniRule~\cite{Bridge2010UniRule}, provide greater
annotation coverage and more regular updates, as annotations are often
transferred from existing annotations.

Database sizes are continuing to expand at an exponential rate,
resulting in a continued and growing reliance on automated curation.
Identification of textual annotation that could be of interest in the
curation process is often based upon biological sequence; sequences
that share properties, such as sequence similarity, are more likely to
share a similar function and attributes. Given a strong sequence
similarity, it is reasonable that annotations may be copied verbatim
between entries, i.e. sentences are subjected to reuse. Therefore,
annotations are often based purely, or in part, on existing
annotations. It is also becoming an increasingly common practice for
manual curators to use existing annotations within their curation
process; either from annotations within the existing database (as is
the case for UniProtKB/Swiss-Prot) or from external databases (e.g.
FlyBase uses UniProtKB as a source). If a database lacks formal
provenance and metadata, it may mean that it is not possible to
identify the original source of an annotation. Given this, the
extracted textual annotation may have also previously been copied from
other entries. Should the original source of a textual annotation be
found to be erroneous, there is no clear way of identifying where it
has propagated to.

A number of studies have explored textual annotation quality (see, for
example,~\cite{Bell12Annotation}), however, very limited work has
explicitly explored textual annotation propagation and its link to
correctness. One such study~\cite{Artamonova05Mining} explores the
usage of association rules to detect possible erroneous annotations.
This study, performed on the Swiss-Prot database, focused primarily on
the annotation within the feature table; free text annotation (those
within the ``CC'' lines) were omitted from the analysis. The reason
for this omission was given as ``[textual annotation] is not easily
machine-parseable''. Unlike structural annotation, textual annotation
is historically developed for human consumption, rather than for
computational interpretation~\cite{Eisenhaber99Evaluation}.
Essentially, this means textual annotations are mostly made up of
free-text English.

Although textual annotation studies are limited, several explore ways
to model propagation of structural annotation
errors~\cite{Gilks02Modeling, Gilks05Percolation, Galperin98Sources}.
Structural annotation sits between nucleotide sequences and textual
annotations; it identifies genomic elements, such as open reading
frames, for a given sequence. This is similar to textual annotation,
in that structural annotation often makes use of sequence data and can
be manually or automatically curated. These
studies~\cite{Gilks02Modeling, Gilks05Percolation, Galperin98Sources}
highlight a number of reasons for structural annotation errors, such
as mis-identification of homology, omissions or typographical
mistakes, concluding that annotation accuracy declines as the database
size increases. Further studies attempt to actually estimate the error
rates in structural annotation. These include an estimated error rate
of between 28\% and 30\% in GOSeqLite~\cite{Jones07Estimating} and
between 33\% and 43\% in
UniProtKB/Swiss-Prot~\cite{Artamonova05Mining}. Therefore, it is
highly plausible that these errors will affect textual annotation, as
acknowledged by~\cite{Gilks02Modeling}.

We hypothesise that sentence reuse is prominent within textual
annotations and a lack of formal provenance has led to inaccuracies in
the annotation space. Within this paper we aim: to quantify sentence
reuse; to investigate patterns of reuse and provenance, through a
novel visualisation technique; and to investigate whether we can use
patterns of propagation to identify erroneous textual annotations,
inconsistent textual annotations or textual annotations with low
confidence.

\section*{Materials and Methods}

\subsection*{The UniProt Knowledgebase (UniProtKB)}

Analyses for this paper focus solely on annotation within
UniProtKB~\cite{UniProt12Organising}. There are a number of reasons
for this. Firstly, UniProtKB consists of two sections: Swiss-Prot,
which is manually curated and reviewed, and TrEMBL, which is automated
and unreviewed. Secondly, the resource is well supported with an
approachable helpdesk and extensive documentation, such as the
UniProtKB user manual~\cite{UPManual}. Finally, UniProt makes
available all past major releases of both Swiss-Prot and TrEMBL, with
the exception of Swiss-Prot versions 1-8 and 10 which were never
archived, in flat file format. UniProtKB also exports the database in
XML format, which includes various levels of evidence
tagging~\cite{UniProt05UniProt}. However, unlike the flat file format,
the XML file format is not available for all versions of the database.
Further, an evidence tag shared between different pieces of data could
be interpreted differently by different users, and thus account for
different sentences. Therefore, the flat file format is used.

Therefore, UniProtKB provides an ideal resource to compare textual
annotation reuse within manually and automatically curated resources
and to investigate its propagation. Since the first version of
Swiss-Prot and TrEMBL a number of key changes in the release process
have occurred. Prior to the formation of the UniProt Consortium, the
releases of Swiss-Prot and TrEMBL were not synchronised; TrEMBL was
released more frequently than Swiss-Prot. In 2004, these releases
became synchronised with UniProtKB initially distinguishing between
major and minor releases until version 15, when this distinction was
abandoned; UniProtKB releases are now on a four week cycle. These
changes make comparison and discussion somewhat challenging, so we
will use the following naming conventions for clarity:

\begin{itemize}
\item {\bf{UniProt}} -- Refers to the UniProt Consortium.
\item {\bf Swiss-Prot} -- Refers to Swiss-Prot database releases prior
  to the formation of the UniProt Consortium.
\item {\bf TrEMBL} -- Refers to TrEMBL database releases prior to the
  formation of the UniProt Consortium.
\item {\bf UniProtKB} -- Refers to the UniProt Knowledgebase,
  including both Swiss-Prot and TrEMBL databases.
\end {itemize}

Where necessary we will explicitly write UniProtKB/Swiss-Prot or
UniProtKB/TrEMBL. This naming scheme allows us to refer to
post-UniProtKB versions of UniProtKB/Swiss-Prot and UniProtKB/TrEMBL
with the same number, starting from version two of UniProtKB, which
was the first major release, containing Swiss-Prot version 44 and
TrEMBL version 27. This numbering scheme continues until version 15,
when subsequent versions follow the format YYYY\_MM, starting from
2010\_01. Complete datasets for historical versions of UniProtKB and
Swiss-Prot are made available by UniProt on their FTP
server\footnote{ftp.uniprot.org/pub/databases/uniprot/}.
Pre-UniProtKB/TrEMBL releases were kindly made available to us by
UniProt.

\subsection*{Sentence Extraction}

Our extraction process has two key parts: a custom made parsing
framework to extract and format the comment lines from UniProtKB
entries and a program to extract the sentences from these formatted
comment lines. The correct extraction of sentences from text is not
straightforward. Given this, we utilised the LingPipe tool
kit~\cite{lingpipe}; a suite of Java libraries for processing text.

A typical sentence within our work is one which contains a group of
words and is terminated with a full stop. However, there are a number
of exceptions to this basic rule, such as abbreviations, which are
especially commonplace in the biomedical domain. The vast majority of
these are handled correctly by LingPipe. However, the structure of
textual annotation in UniProtKB can mean topic blocks or lists are not
terminated with a full stop. In cases such as these, we count these as
sentences. Specifically, our extraction process, as summarised in
Figure~\ref{fig:extractionProcess}, involves:

 \begin{figure}[!ht]
 \begin{center}
 \includegraphics[width=0.45\textwidth]{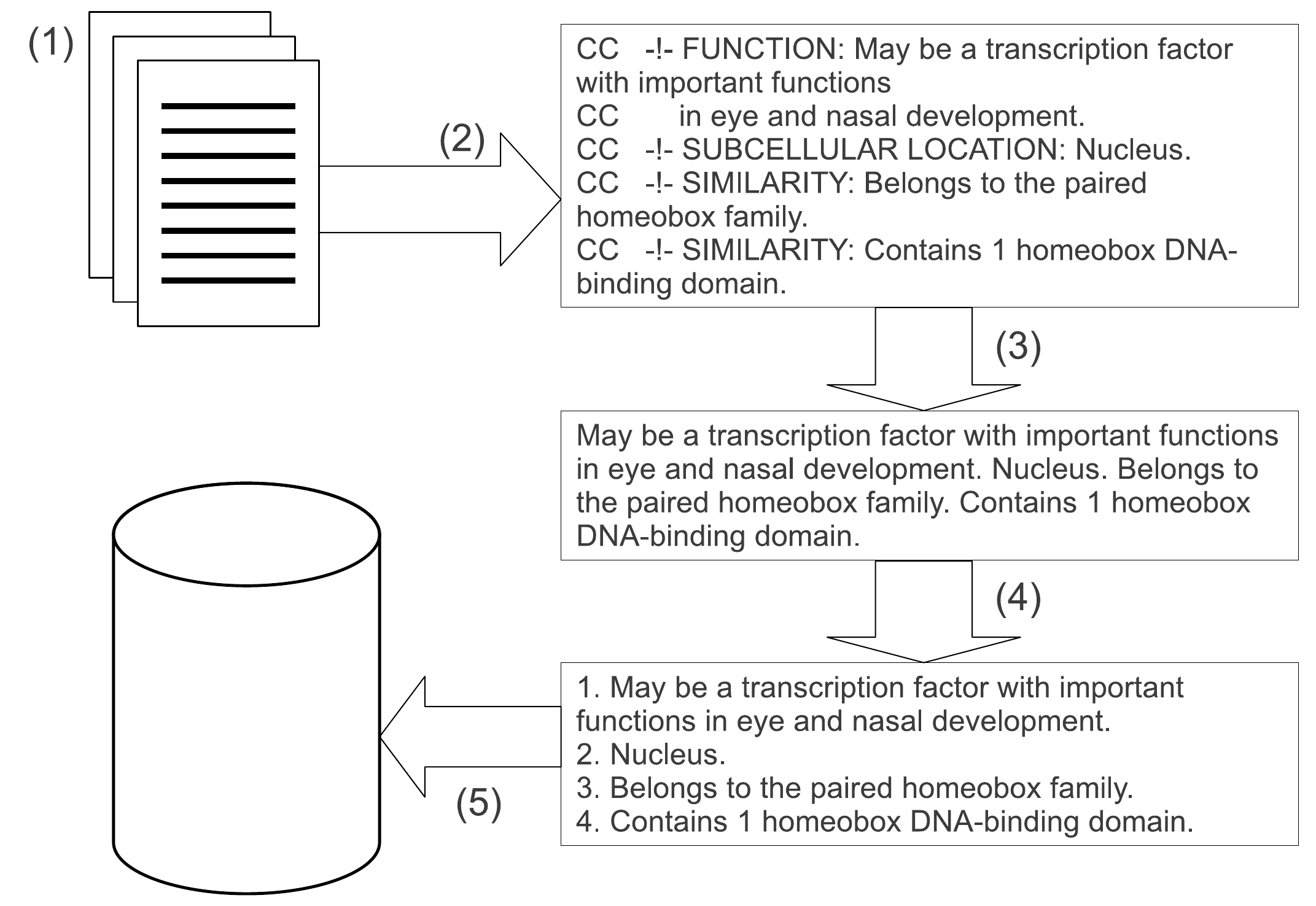}
 \end{center}
 \caption{{\bf Outline view of the data extraction process.} (1)
   Initially we download a complete dataset for a given database
   version in flat file format. (2) We then extract the comment lines
   (lines beginning with `CC', the comment indicator). (3) We remove
   comment blocks and properties (as defined in the UniProtKB
   manual~\cite{UPManual}) and the `CC' identifier. (4) Sentences are
   then extracted, using LingPipe. (5) Finally, all of the identified
   sentences are added to the MySQL database.}
 \label{fig:extractionProcess}
 \end{figure} 

\begin{enumerate}
\item Downloading and extracting complete datasets from historical
  versions of UniProtKB in flat file format from the UniProt FTP
  server\footnotemark[\value{footnote}].
\item Extracting comment lines from these flat files using the Java
  framework created to handle the UniProtKB flat file structure, as detailed
  in the UniProtKB user manual~\cite{UPManual}.
\item Removing topic headings, the ``CC'' identifier and copyright and
  licence statements. Over time annotations in UniProtKB have become more
  structured with the addition of topic headings (e.g. ``subcellular
  location'' and ``function'') in the comments lines, which were
  removed to maintain sentence integrity.
\item Extracting a list of all the sentences from each entries comment
  lines using LingPipe.
\item Storing extracted sentences in a MySQL database, stating the
  entry it appears in and for which database version.
\end{enumerate}

To ensure that annotation data was correctly extracted from UniProtKB,
a number of checks were performed. These checks mostly involved making
use of the UniSave tool~\cite{Leinonen06UniSave}, made available by
UniProt. UniSave allows the differences within an entry to be compared
between two different versions. Making use of this tool we were able
to manually check that sentences were correctly parsed for a random
selection of entries and versions.

Once we extracted all of the sentences from every entry within a given
database version, we had a set of sentences which we refer to as the
\emph{total} number of sentences within a database version. This set
of sentences is redundant, as a number of sentences will occur
multiple times within the set. Taking each sentence from this set only
once (i.e. extract the distinct sentences) results in a set of
non-redundant \emph{unique} sentences. Finally, within a set of unique
sentences, some sentences occur only a single time within a database
version; that is they are \emph{singleton} sentences. We can summarise
these definitions as:

\begin{itemize}
\item \textbf{Total sentences} -- A redundant set of all sentences in
  a database version.
\item \textbf{Unique sentences} -- A non-redundant set of all
  sentences in a database version.
\item \textbf{Singleton sentences} -- A set of sentences that occur
  only a single time within an entire database version.
\end{itemize}

\section*{Results}

\subsection*{How heavily reused are sentences in UniProtKB?}

\begin{figure*}[!ht]
\begin{center}
\includegraphics[width=6in]{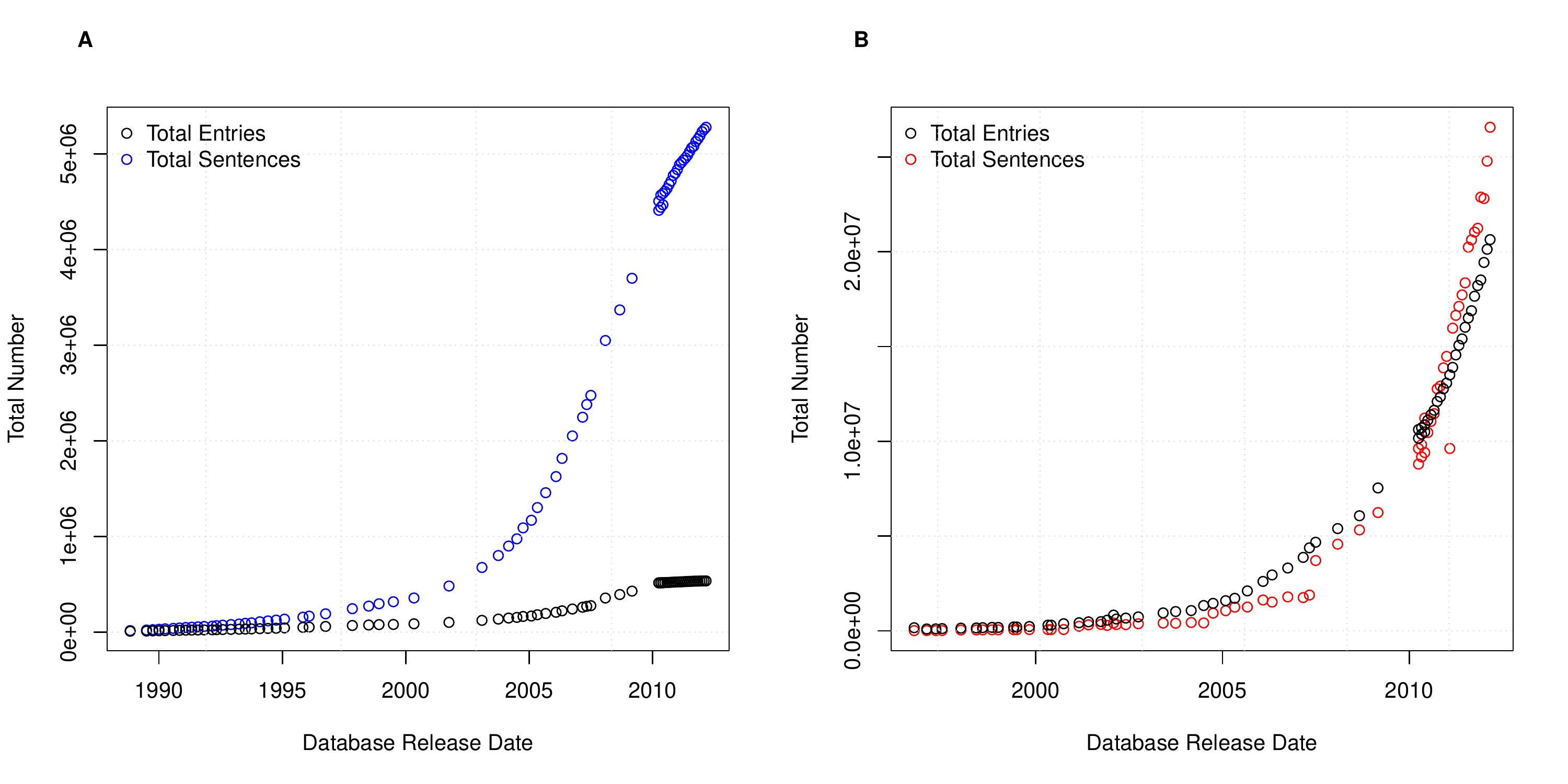}
\end{center}
\caption{ {\bf Total sentences and entries.} The total number of
  sentences and entries in (A) Swiss-Prot and (B) TrEMBL.}
 \label{fig:entriesVsSentences}
 \end{figure*}

The curation process implemented by UniProtKB~\cite{Magrane11UniProt}
means that sentences will be reused. To understand the amount of
sentence reuse over time, we initially analyse the total number of
sentences that are used within each version of Swiss-Prot and TrEMBL.
These results, as shown in Figure~\ref{fig:entriesVsSentences},
clearly show that the total number of sentences is growing rapidly.
Whilst the growth for Swiss-Prot shows a relatively regular
progression, TrEMBL has a somewhat more irregular and disjointed
growth. Figure~\ref{fig:entriesVsSentences} also shows the number of
entries within UniProtKB over time. This figure shows that the growth
of sentences within TrEMBL generally follows the growth of the
database, whilst the growth of sentences in Swiss-Prot is much slower
than the growth of entries.

\begin{figure}[!ht]
 \begin{center}
 \includegraphics[width=0.45\textwidth]{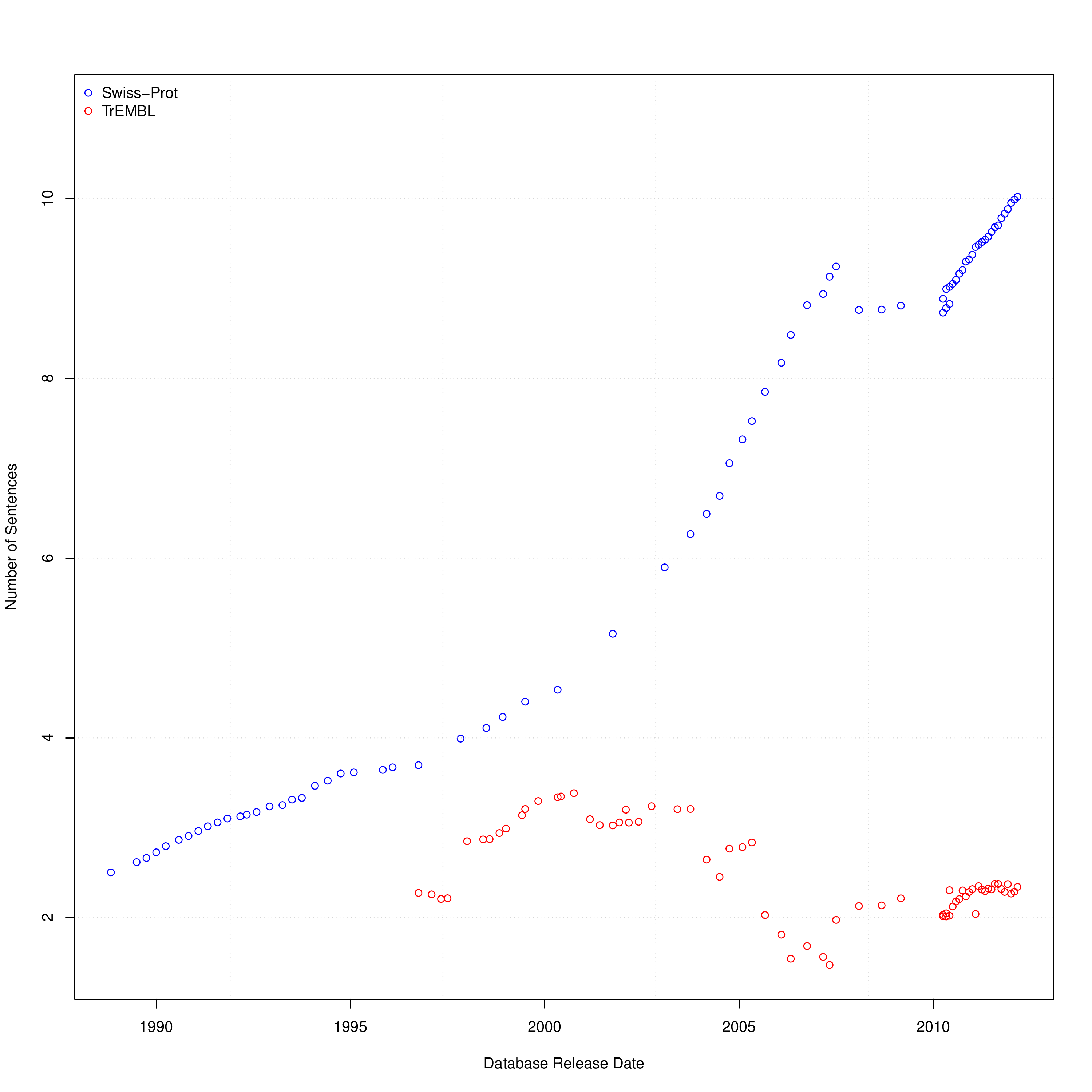}
 \end{center}
 \caption{{\bf Average sentences per entry.} The number of sentences
   that appear, on average, in each entry in TrEMBL and Swiss-Prot
   (i.e. the total number of sentences divided by the total number of
   entries).}
 \label{fig:averageSentencesEntry}
 \end{figure}

Both the size of the database and the number of sentences is
increasing over time. How, then, does this impact the way sentences
are re-used and distributed within the database? We can gain an
insight into the re-use by analysing the average number of sentences
per entry, as shown in Figure~\ref{fig:averageSentencesEntry}. For
this calculation, only entries containing textual annotation were
considered. Figure~\ref{fig:averageSentencesEntry} shows that, over
time, entries within Swiss-Prot have an increasing number of sentences
in their annotations. Over a twenty year period, Swiss-Prot has seen
the number of sentences within the textual annotation of an entry
increase fivefold, to the current average of around ten. Conversely,
TrEMBL has seen fluctuations over time, but typically remained at an
average of around two sentences per entry.

\begin{figure*}[!ht]
 \begin{center}
 \includegraphics[width=\textwidth]{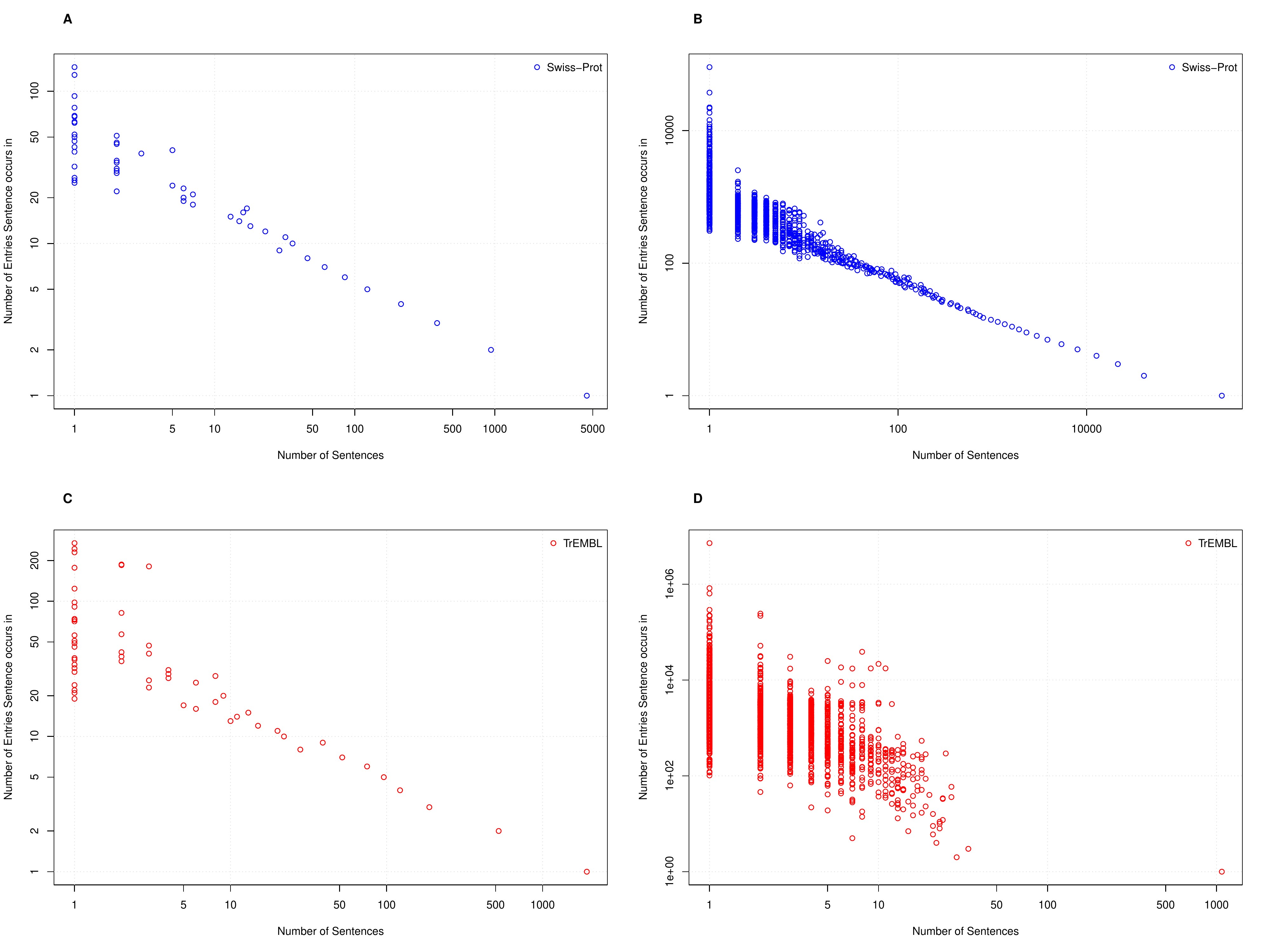}
  \end{center}
  \caption{{\bf Showing the distribution of sentence reuse through
      Swiss-Prot and TrEMBL.} (A) Swiss-Prot Version 9 (B)
    UniProtKB/Swiss-Prot Version 2012\_05 (C) TrEMBL Version 1 (D)
    UniProtKB/TrEMBL Version 2012\_05. As an example, in Figure A, the
    bottom right point states that there is $\sim$5000 sentences that
    occur a single time, whilst the top-left-most point indicates that
    there is one sentence that occurs $\sim$125 times.}
 \label{fig:sentenceDistribution}
 \end{figure*}

To complement this reuse analysis, we can also analyse how sentences
are distributed throughout UniProtKB, as shown by
Figure~\ref{fig:sentenceDistribution}. This shows the distribution is
much more even in Swiss-Prot than TrEMBL, whilst again highlighting
the increasing levels of reuse over time.

\begin{figure*}[!ht]
\begin{center}
\includegraphics[width=6in]{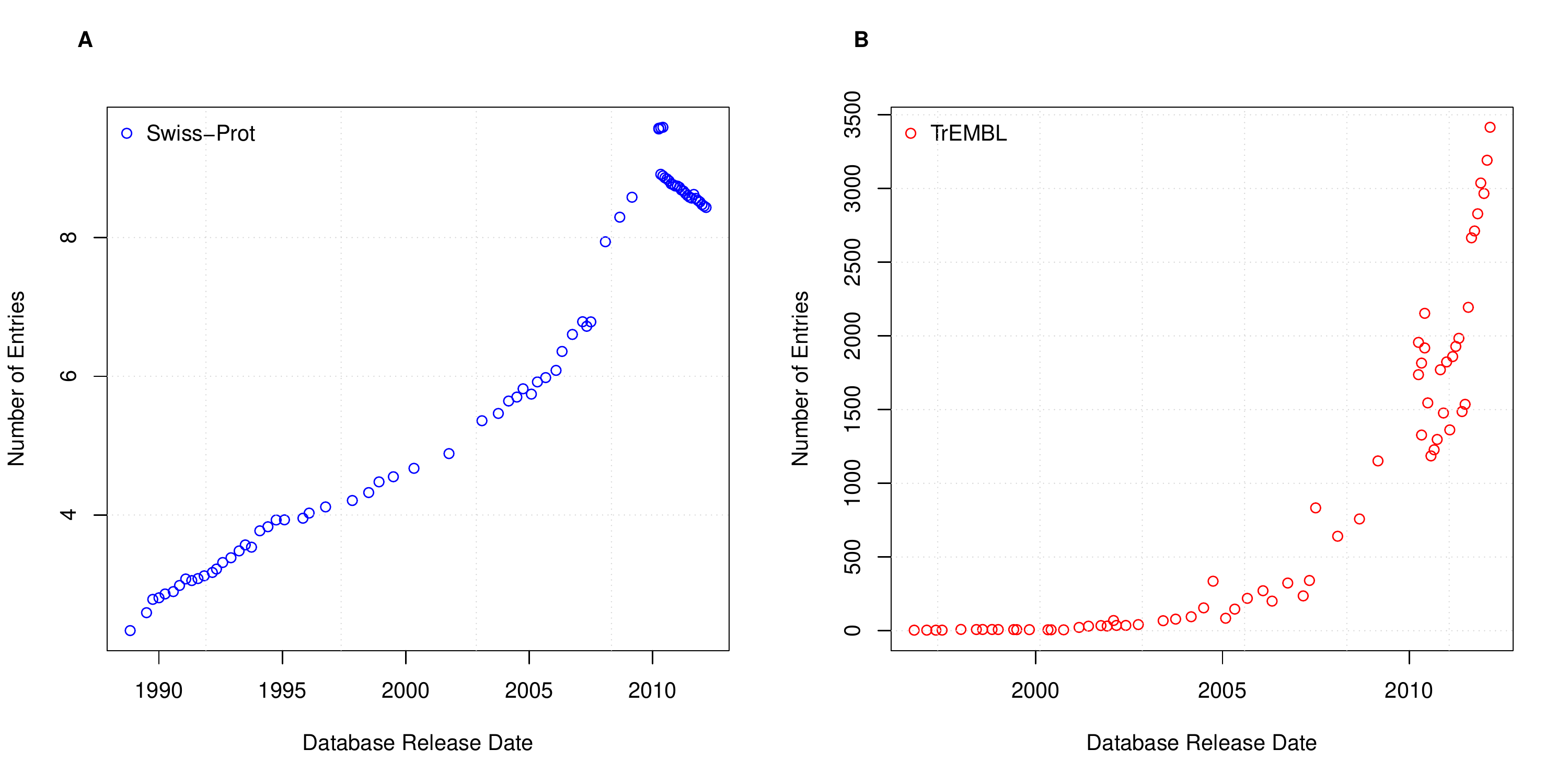}
\end{center}
\caption{ {\bf Average Entries Per Sentence.}  Graph showing the
  average number of entries that each sentence appears in for (A)
  Swiss-Prot and (B) TrEMBL.  }
\label{fig:averageEntriesPerSentence}
\end{figure*}

The increase in the number of sentences in the textual annotation per
entry over time fits with one of the goals of UniProtKB, which is to
attach as much information as possible to each protein
entry~\cite{Apweiler04UniProt}. A significant amount of this increase
is through sentence reuse. We can see this by considering the number
of entries that each unique sentence occurs in;
Figure~\ref{fig:averageEntriesPerSentence} shows that the average
number of entries where a particular sentence appears is generally
increasing for Swiss-Prot and TrEMBL, to a current average of about 9
and 3500 respectively; interestingly later versions of Swiss-Prot are
starting to show a decline in reuse, which coincides with the change
in release cycle of UniProtKB.

These results suggest that whilst total textual annotation is
increasing for entries on average, it is driven by sentence
reuse. Another factor affecting the amount of sentence reuse could be
UniProtKB attempting to reduce the number of entries that remain
without any textual annotation.

\begin{figure*}[!ht]
\begin{center}
 \includegraphics[width=6in]{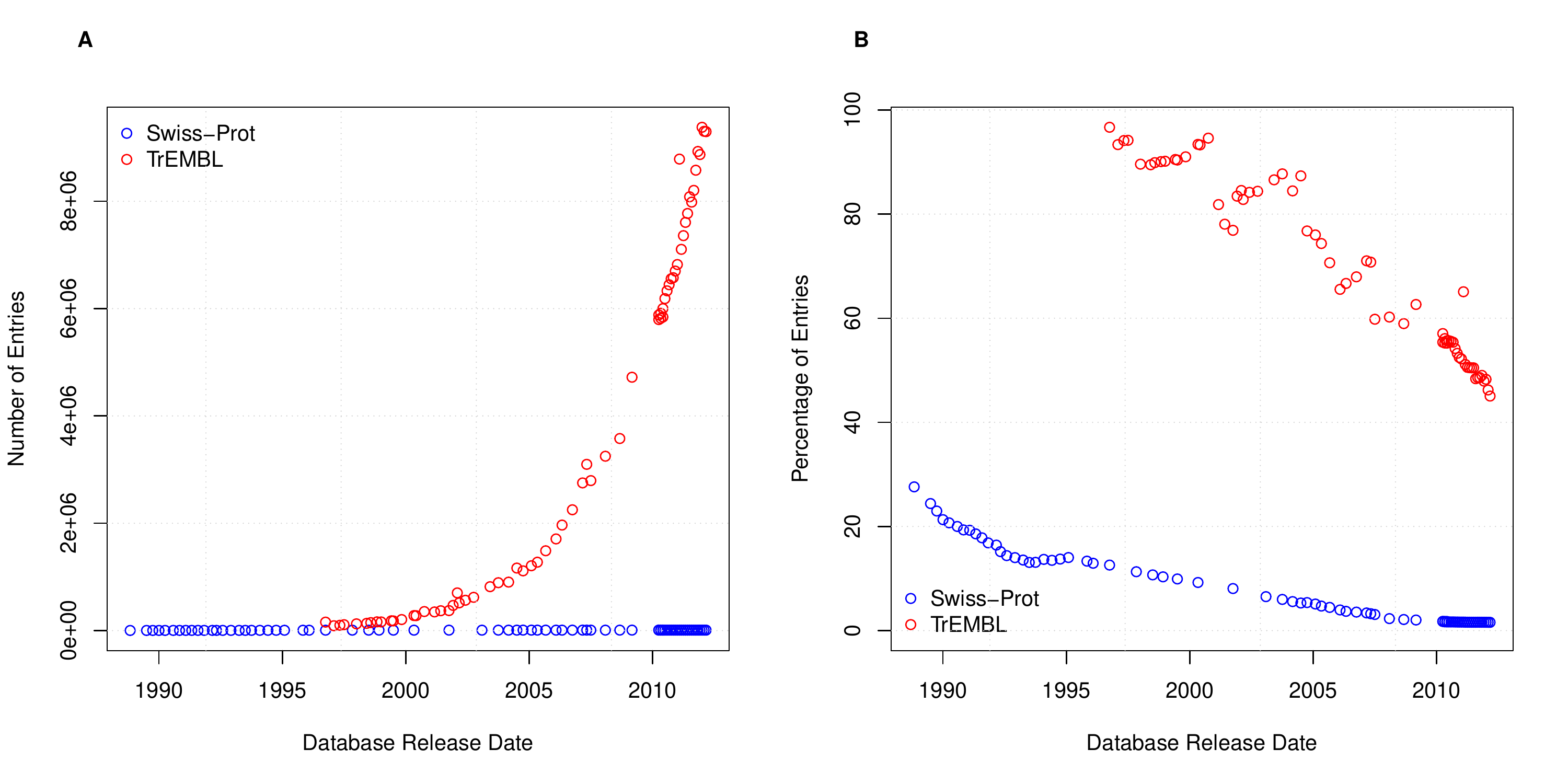}
  \end{center}
  \caption{ {\bf Entries without annotation.} (A) Number of entries in
    TrEMBL and Swiss-Prot without any annotation, and (B) the
    percentage of entries without any annotation}
 \label{fig:entriesWithoutAnn}
 \end{figure*}

We show the number of entries without any textual annotation in
Figure~\ref{fig:entriesWithoutAnn} (A), and the overall percentage in
Figure~\ref{fig:entriesWithoutAnn} (B). Over time the overall
percentage of these entries is decreasing; only around 1.5\% of
entries in the latest version of Swiss-Prot contain no textual
annotation, compared to 45\% of entries in the latest version of
TrEMBL. Both of these show significant improvements over time --
initially Swiss-Prot had 27.6\% of entries without any textual
annotation in 1988 compared to TrEMBL which had 96.7\% in 1996.

Therefore, we conclude that, in addition to the increase of the
overall database size, the percentage of entries with annotation is
increasing; these two factors both contribute to the increasing
reuse of sentences. 

These results suggest that the amount of textual annotation is
increasing due to an increase in sentence reuse. We therefore want to
abstract from the overall reuse and ask: how is the number of unique
sentences changing over time?

\begin{figure*}[!ht]
 \begin{center}
 \includegraphics[width=6in]{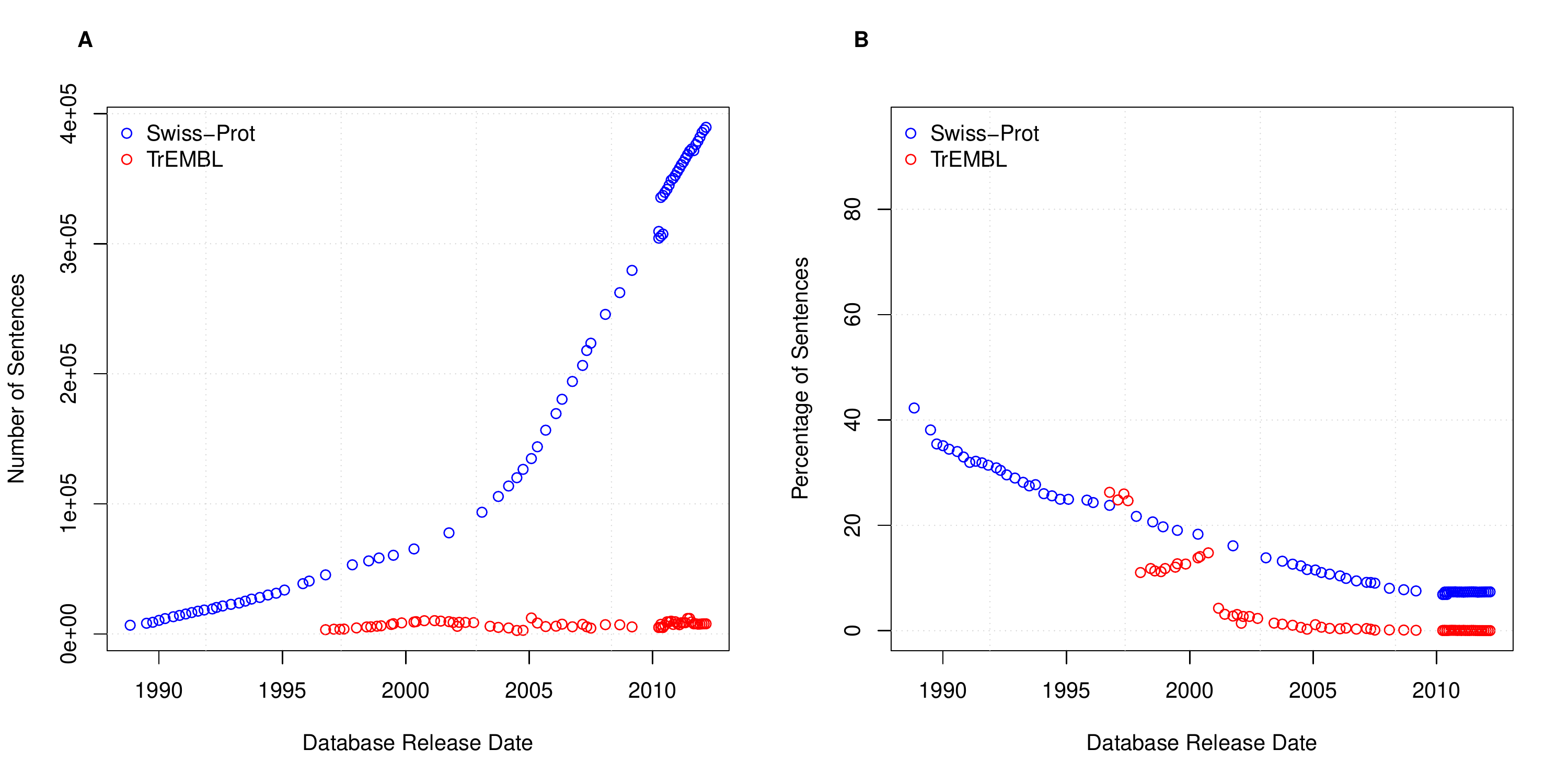}
  \end{center}
  \caption{ {\bf Unique sentences.} (A) The number of unique sentences
    in Swiss-Prot and TrEMBL and (B) the percentage of unique
    sentences in Swiss-Prot and TrEMBL.}
 \label{fig:uniqueStats}
 \end{figure*}

Figure~\ref{fig:uniqueStats} (A) shows the level of unique sentences
within both TrEMBL and Swiss-Prot. From this, it is immediately clear
that sentences are much more heavily reused within TrEMBL than
Swiss-Prot. To further illustrate this, in
Figure~\ref{fig:uniqueStats} (B) we show the percentage of unique
sentences for each database version of Swiss-Prot and TrEMBL. This
graph shows a steady decline in both Swiss-Prot and TrEMBL, providing
further evidence that sentence reuse in both databases is on the
rise. For example, within UniProtKB/TrEMBL Version 2012\_05 there are
approximately 22 million entries, containing approximately 26.7
million sentences, 8131 of which are unique; i.e. the entire TrEMBL
sentence corpus is made up of only $0.03\%$ sentences.

\begin{figure*}[!ht]
\begin{center}
 \includegraphics[width=6in]{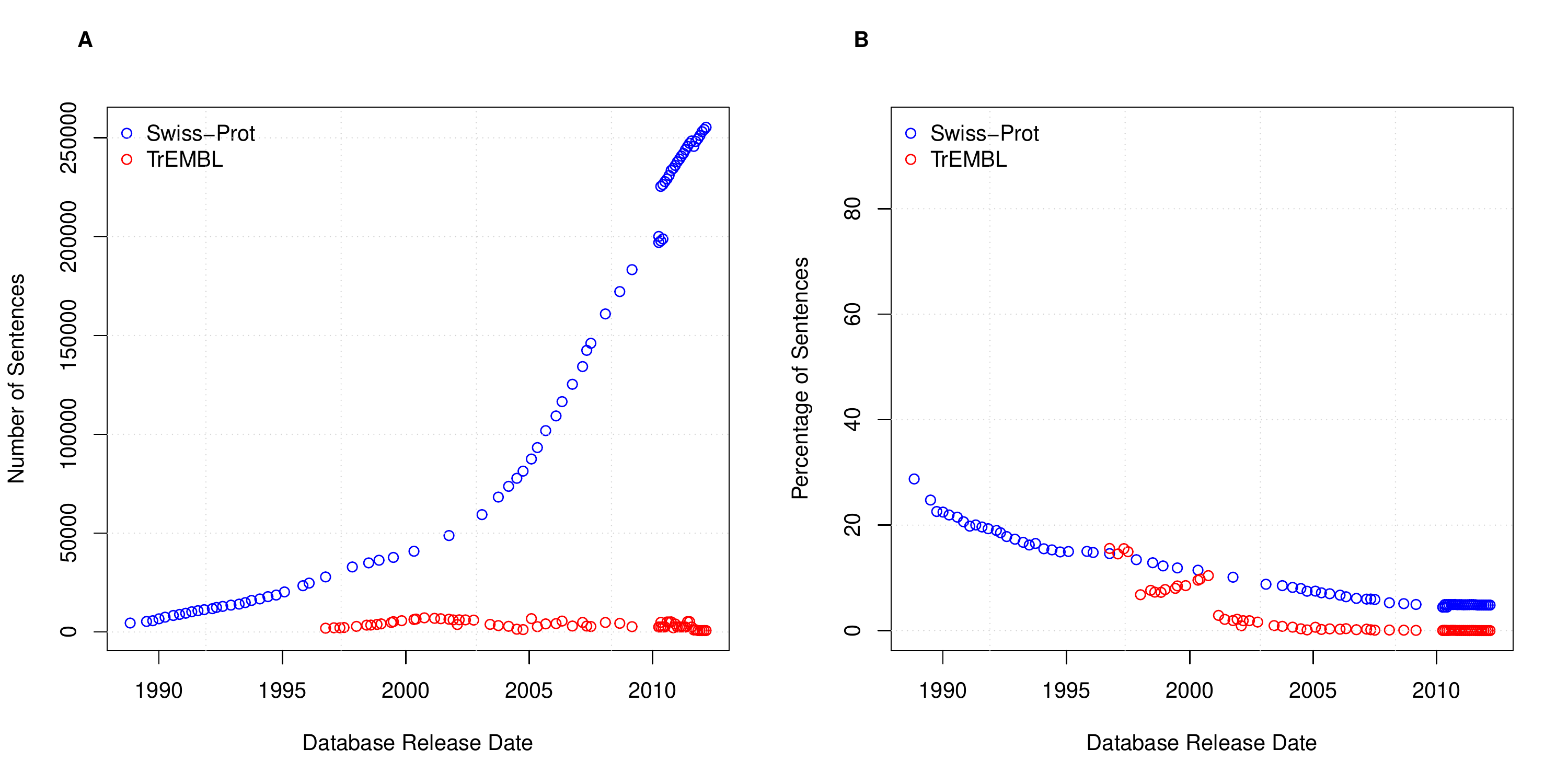}
   \end{center}
   \caption{ {\bf Singleton sentences.} The number of singleton
     sentences in Swiss-Prot and TrEMBL and (B) the percentage of
     singleton sentences in Swiss-Prot and TrEMBL.}
 \label{fig:singletonStats}
 \end{figure*}

A special case of the unique sentence is the singleton sentence, that
is a sentence which occurs once, and only once, within a database
version. The number of singleton sentences is shown in
Figure~\ref{fig:singletonStats} (A) with the percentage shown in
Figure~\ref{fig:singletonStats} (B). Figures~\ref{fig:singletonStats}
and~\ref{fig:uniqueStats} show both singleton and unique sentences
follow an almost identical pattern.

In all cases we see that while absolute numbers are increasing,
percentages are decreasing. We see this for both unique sentences
($389,558$ and $7760$ ($\sim$7\% and $\sim$0.03\%) in the latest
versions of Swiss-Prot and TrEMBL, respectively) and singleton
sentences ($255,349$ and $735$ ($\sim$5\% and $\sim$0.003\%) in the
latest versions of Swiss-Prot and TrEMBL, respectively). Therefore, in
conclusion, this shows that reuse between records is increasing in
both Swiss-Prot and TrEMBL, and this trend appears set to continue.

Whilst we are able to quantify this reuse, we are currently unable to
analyse and depict the reuse of individual sentences; we would like to
analyse and explore how an individual sentence is propagated through
the database. For this analysis to be performed we need to identify
possible visualisation approaches, as explored in the following
section.

\subsection*{How can we visualise sentence propagation?}

We have shown that sentence reuse in UniProtKB is both common and
increasing. Therefore, given the scale of this data, we explore the
usage of \emph{visualisation}. By visualising sentence reuse across
entries and over time, we may be able to better understand annotation
propagation and infer provenance. From this, it is possible that
patterns demonstrating interesting traits in the underlying data may
emerge and be identified. Therefore, we wish to explore how we can
visualise this data and ask: how can we clearly represent the flow of
annotation through the database?

A number of approaches to visualising large datasets were considered.
One such approach is to model the relationship between sentences and
entries as a graph. Using a tool, such as
Cytoscape~\cite{Smoot11Cytoscape}, we can easily model sentences
occurring within entries (and vice versa). However, our experience
with this approach suggests that it is troublesome to model change
over time and manual intervention is often required to ensure nodes
are organised in a correct and meaningful manner. Other similar
approaches, such as Sankey diagrams, were not utilised as we cannot
determine the exact source and flow of an annotation between each
individual entry.

One approach which produces a visualisation similar to our
requirements is the history flow tool~\cite{Viegas04Studying}. This
tool was developed to allow visualisation of relationships between
multiple versions of a wiki. Therefore, it aims to clearly depict the
change in sentences, and their order, in a document over time with the
ability to attribute each change to a given author. The authors
demonstrated this visualisation with an exploratory analysis of
Wikipedia, revealing complex patterns of cooperation and conflict
between Wikipedia authors. However, using the history flow tool to
visualise the flow of individual sentences in UniProtKB is not ideal;
crucially, the tool cannot clearly represent the data due to the
disjoint nature of early Swiss-Prot and TrEMBL releases.

\begin{figure}[!ht]
\begin{center}
\includegraphics[width=0.45\textwidth]{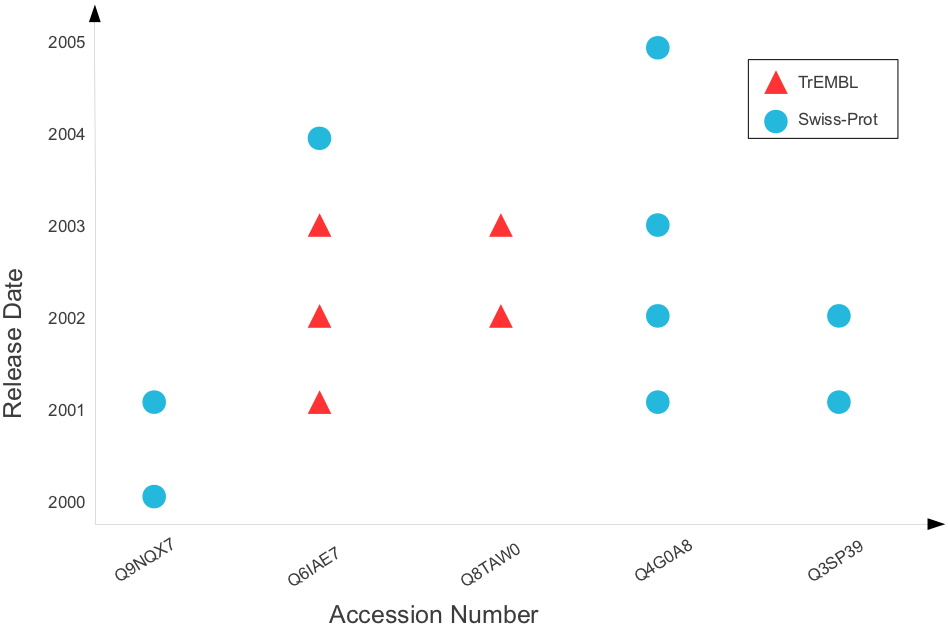}
  \end{center}
  \caption{{\bf Manual illustration showing how the propagation of a
      single sentence could be visualised.} Accession numbers are
    shown on the X-Axis, with database release dates shown on the
    Y-Axis. A point on the graph represents that the sentence occurs
    in an entry within a given database version. For example, the
    bottom left point shows that the sentence occurs in accession
    entry Q9NQX7 for Swiss-Prot in 2000 -- this sentence remains in
    Q9NQX7 for one more version; it is removed in the following
    version (in 2002).}
\label{fig:sentIll}
\end{figure}

Given these issues, we look at creating a visualisation view of the
annotation space that overcomes these restrictions, whilst also aiming
to make the visualisation as intuitive as possible. We outline a
visualisation approach, as manually illustrated in
Figure~\ref{fig:sentIll}, that is somewhat similar to a regular
scattergraph plot and draws upon the strengths from the history flow
tool. This approach allows propagation to be visualised whilst also
remaining intuitive; we show each accession the sentence occurs within
along the X-axis, with the Y-Axis showing the release date for the
corresponding database versions. Therefore, a point on this graph
represents the sentence occurring within an accession for a given
database version, where a red point represents the sentence being in a
TrEMBL entry, and a blue point represents a Swiss-Prot entry.

This approach quickly becomes cumbersome as the amount of data
increases, specifically making it difficult to explore and examine
individual data points. Additionally, there can be several TrEMBL
releases for each Swiss-Prot release.  This makes it appear that the
sentence is constantly being removed and re-added; i.e. it exhibits
\emph{striping}. This striping is due to early releases of TrEMBL and
Swiss-Prot being unsynchronised. One possibility to overcome the issue
of striping is binning. However, this would lose a major level of
granularity, as we would have to bin for every six months to cover all
Swiss-Prot releases.

\begin{figure*}[!ht]
{\centering
 \includegraphics[width=\textwidth]{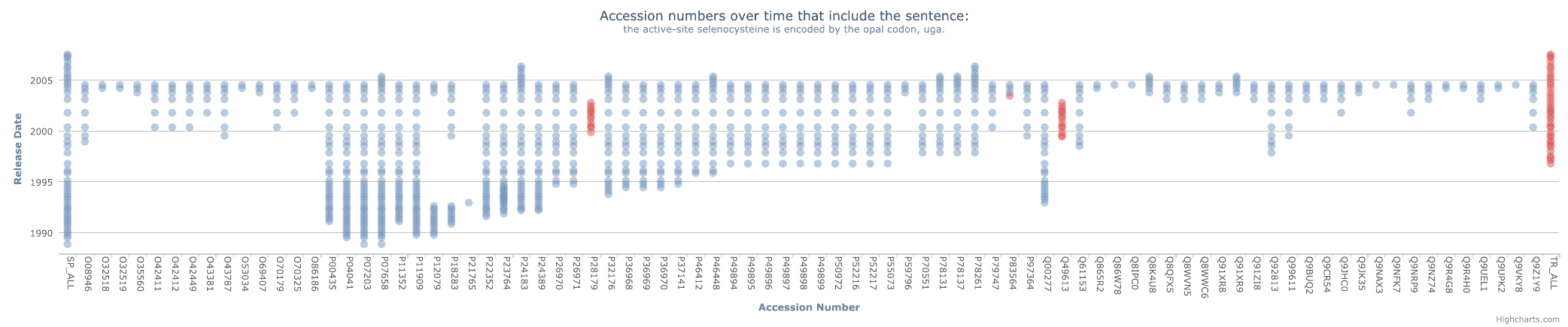}
 }
 \caption{{\bf Visualising sentence propagation.} Visualising the
   propagation of the sentence ``the active-site selenocysteine is
   encoded by the opal codon, uga.'' through the database, with all
   possible versions of Swiss-Prot and TrEMBL within this range shown
   at either end of the graph.}
 \label{fig:interGraph}
 \end{figure*}

\begin{figure*}[!ht]
\begin{center}
\includegraphics[width=6in]{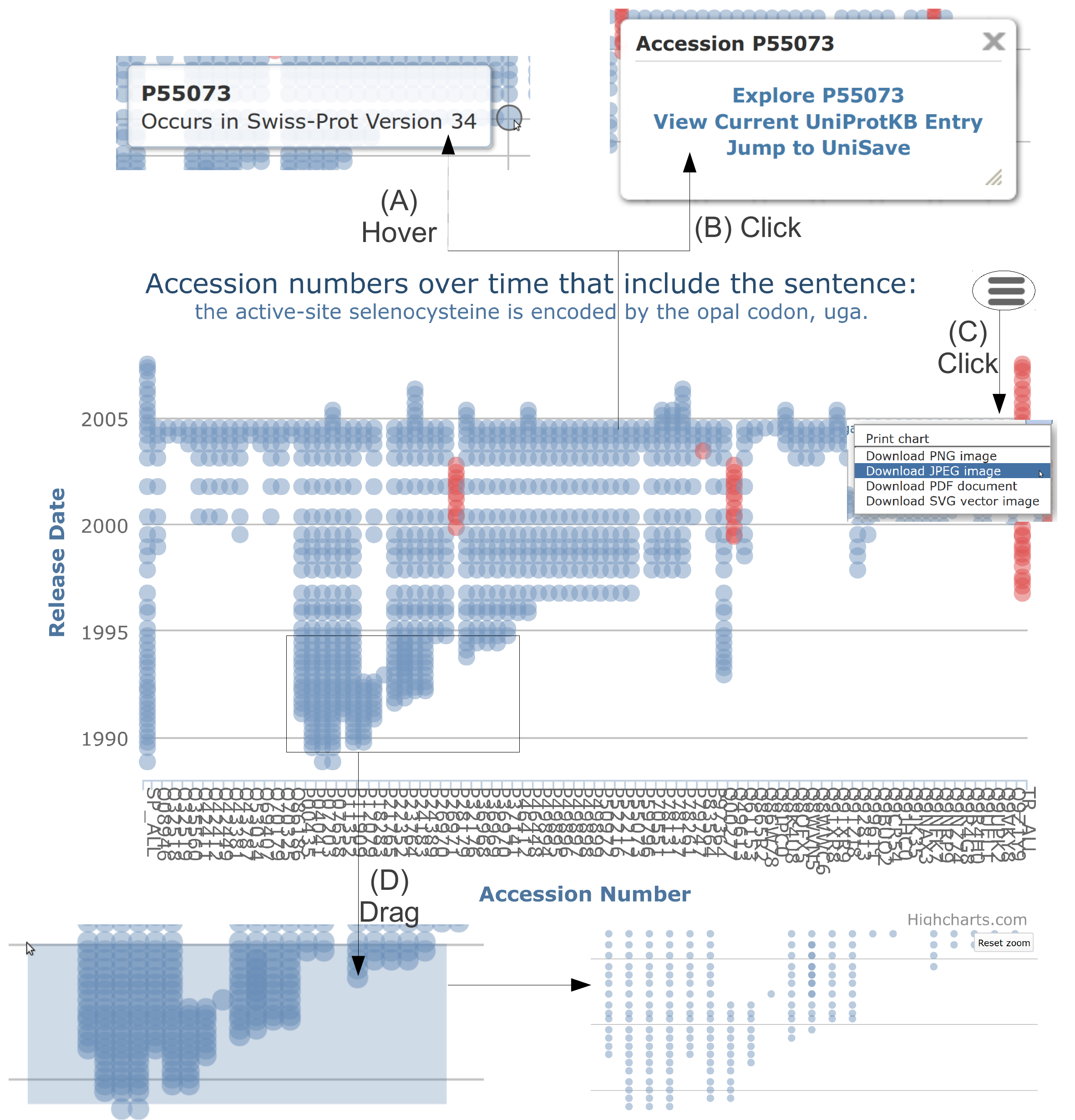}
\end{center}
\caption{ {\bf Illustrating the key interactive features provided by
    Highchart graphs.} (A) Hovering over a point indicates the
  corresponding accession number and database version. (B) Clicking on
  a point provides links to the UniProt entry and further
  information. (C) Each graph can be printed and exported into a
  variety of image formats. (D) The ability to zoom into a section of
  a graph; this can be achieved by left-clicking and dragging a
  desired area.}
 \label{fig:highcharts}
 \end{figure*}

To overcome these issues, we explored generating graphs using an
interactive framework. The resulting graph, for the sentence ``the
active-site selenocysteine is encoded by the opal codon, uga.'' is
show in Figure~\ref{fig:interGraph}. These graphs make use of
Highcharts, an interactive charting library in
JavaScript\footnote{http://www.highcharts.com/}. This approach
provides an interactive web-based chart option, that can be easily
generated for any sentence. Further, we have the ability to zoom into
dense graphs, hover over a point to clearly see the entry and
corresponding database version and export graphs (i.e. save to
file). These features, as illustrated in Figure~\ref{fig:highcharts},
allow us to overcome the issues caused by dense graphs. Additionally,
we show the release points for Swiss-Prot down the left side and
TrEMBL down the right, thus making it clearer when a point is missing
and further alleviating the striping issues.

Within the UniProtKB database, it is relatively common for entries to
become merged. When a merge occurs, the entry has a single primary
accession, with the merged entries becoming secondary accessions.
Within our graphs, we show an entry by all of its accessions; not
doing so would be misleading, as it would appear that an annotation
has been removed when, in reality, the entry has been merged.

The development of this visualisation approach allows us to
investigate further how individual sentences have been used within
UniProtKB, and how they move between different entries over time. We
next discuss how we have used this visualisation strategy.

\subsection*{Exploring the annotation space: Can provenance be
  identified?}

We have shown that sentence reuse is frequent within UniProtKB and is
increasing as UniProtKB matures. With the development of a
visualisation technique, we are now able to visualise sentence
propagation, with the specific aim of investigating the provenance of
annotations.

Taking as an initial example the sentence ``the active-site
selenocysteine is encoded by the opal codon, uga.'', we show the
visualisation of its propagation in Figure~\ref{fig:interGraph}. This
graph shows that the sentence initially occurs in two entries in
Swiss-Prot Version 9; P07658 and P07203 (the leftmost point, SP\_ALL,
is for illustration and used to alleviate striping, as previously
discussed). In this particular instance, the provenance is between two
entries -- we cannot trace this further back as Swiss-Prot versions
1-8 and 10 are missing. Additionally, our level of granularity shown
within these graphs is at major release level. Therefore, it is
possible that a sentence will appear to originate in two or more
entries within a single database version, when a distinction between
them could be made at the minor release level. Minor release data was
not parsed as it is only accessible through UniSave.

\begin{figure}[!ht]
{\centering
 \includegraphics[width=0.45\textwidth]{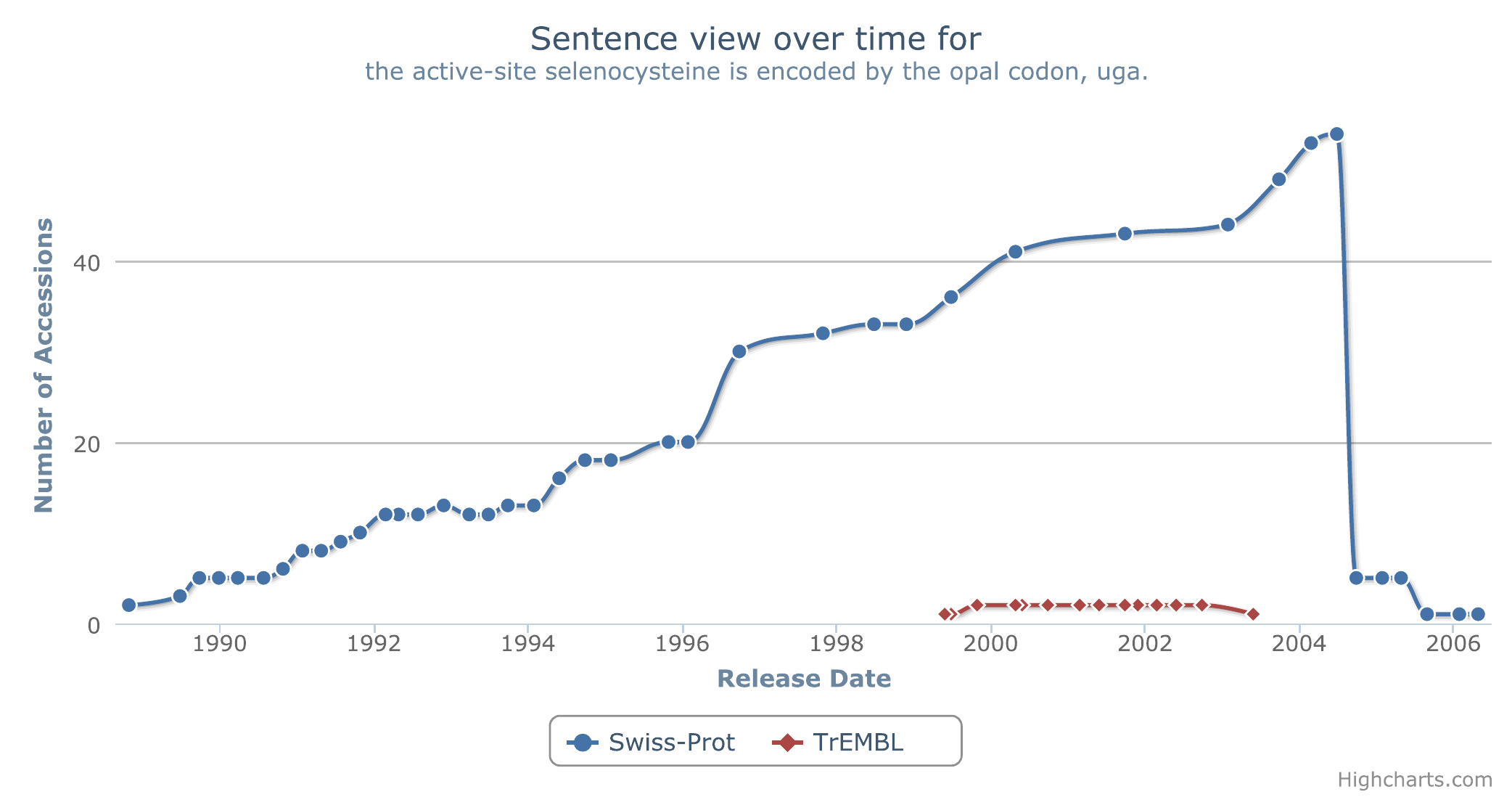}
    }
    \caption{{\bf Visualising sentence occurrences.} The number of
      UniProtKB entries that the sentence ``the active-site
      selenocysteine is encoded by the opal codon, uga.''  appears in
      over time}
 \label{fig:selenoOccur}
 \end{figure}

Having identified that the sentence originated in two entries, we can
also show that, at its peak, it was most commonly seen in Swiss-Prot
Version 44, where it was found in a total of 54 Swiss-Prot entries, as
illustrated in Figure~\ref{fig:selenoOccur}. In total, the sentence
appeared in 84 unique entries within UniProtKB until its removal. 

The removal of this sentence was due to formatting changes within
UniProtKB. Essentially, a change in the UniProtKB protocol meant that
information about selenocysteine encoding moved from the textual
annotation to the feature table of UniProtKB entries. Such technical
changes are inevitable given the age of UniProtKB, which is constantly
evolving to meet the requirements of new and updated developments.
Such technical changes will inevitably impact a number of sentences.
However, in this particular case we have identified a sentence that
should have been replaced in all entries.

\begin{figure}[!ht]
{\centering
 \includegraphics[width=0.45\textwidth]{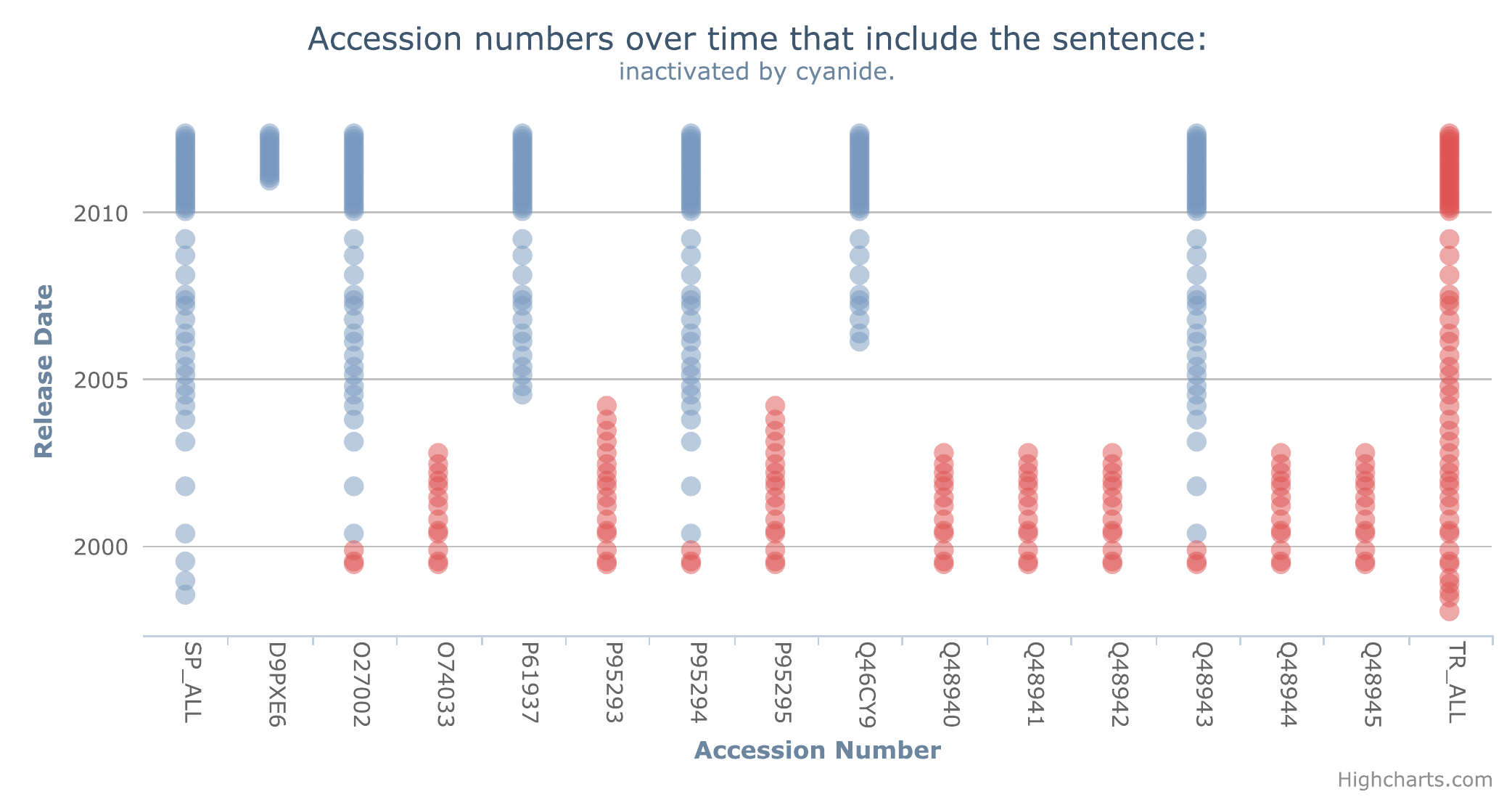}
    }
    \caption{{\bf Visualising a sentence originating in TrEMBL.} An
      example of a sentence (``inactivated by cyanide.'') that
      originates in TrEMBL, but ends up in Swiss-Prot. In this case, a
      number of the TrEMBL entries are merged into Swiss-Prot.}
 \label{fig:TrEMBL_First}
 \end{figure}

By analysing the flow of the sentence throughout UniProtKB in
Figure~\ref{fig:interGraph}, we notice a number of interesting
\emph{propagation patterns}:

\begin{itemize}

\item \textbf{Missing origin} -- The sentence is removed from the
  entries it first originated in, yet still remains in a number of
  other entries in the database after this point.

\item \textbf{Reappearing entry} -- In two entries (P18283 and P12079)
  the sentence is removed, with the sentence actually being re-added
  to each of these entries after a number of versions have elapsed.

\item \textbf{Transient appearance} -- In a number of entries, such as
  P21765, the sentence only appears for a single version. It is
  removed from the subsequent release.

\item \textbf{Originating in TrEMBL} -- Although not shown in
  Figure~\ref{fig:interGraph}, there are cases where a sentence
  originates in TrEMBL, before being propagated into Swiss-Prot entries. An
  example of this pattern is shown in Figure~\ref{fig:TrEMBL_First}.

\end{itemize}

It is clear that by visualising sentences in this manner we are able
to detect provenance. Moreover, inspection of these graphs has led to
the discovery of a set of propagation patterns. These patterns are
unexpected; why, for instance, should a sentence appear in only a
single version of UniProtKB, or should a sentence disappear in an
originating entry, but remain in an apparently derived entry? Given
this, we now wish to examine these patterns further, exploring how
frequently each pattern occurs within the database and what quality
information can be drawn from them.

\subsection*{Exploring the annotation space: Analysing propagation
  patterns}

In the previous section, four propagation patterns were identified
through the examination of sentences such as ``the active-site
selenocysteine is encoded by the opal codon, uga.''. If these patterns
are of analytical value, then we would suspect a significant number of
sentences adhering to each pattern to exist within the database. To
obtain a list of these sentences, we used a series of set operations.
For example, to identify sentences which follow the missing origin
pattern, we take two sets: the entry, or entries, that the sentence
first occurred within and the entries that the sentence last occurred
within. If taking the intersection of these sets results in an empty
set, then we have identified a sentence that is missing its root
origin. Using this approach, an algorithm was created to allow the
automated identification and extraction of sentences for each of the
identified patterns.

Many sentences which exhibit each pattern were extracted from the
UniProtKB database, with these results summarised in
Table~\ref{tab:patternSummary}. In total, over $85,000$ sentences
follow at least one of the identified patterns, with over $35,000$
sentences remaining in the latest version of UniProtKB; in other
words, approximately 9\% of the unique sentences in UniProtKB Version
2012\_05 follow one of the identified patterns.  Amongst these
patterns, transient sentences are the most prominent, accounting for
approximately $75\%$ of the sentences following one of these patterns.

\begin{table*}[!ht]
\centering
\caption{
\bf{Number of identified propagation patterns}}
\begin{tabular}{|l|l|l|}
  \hline
  \textbf{Pattern} & \textbf{Number of sentences} & \textbf{Number in just UniProtKB Version
    2012\_05} \\
  \hline 
  Missing Origin & 8355 & 3835 \\
  \hline
  Reappearing Entry & $15,587$ & 7011 \\
  \hline
  Transient appearance & $68,042$ & $25,582$ \\
  \hline
  Originating in TrEMBL & 8649 & 5330 \\
  \hline
\end{tabular}
\begin{flushleft}The number of sentences that adhere to each pattern, for all
  versions of UniProtKB and those just in the latest version of
  UniProtKB. To place these results in context, there have been a
  total of $611,080$ unique sentences, with $394,233$ unique sentences
  being in UniProtKB Version 2012\_05.
\end{flushleft}
\label{tab:patternSummary}
\end{table*}

We have defined a transient sentence as one which is only present in
an entry for a single database release before removal. By revisiting
Figure~\ref{fig:interGraph}, it can be seen that there are six
instances for the given sentence where this occurs. Five of these
cases occurred in Swiss-Prot version 44, when the majority of
sentences were removed. The other case only occurs in entry P21765 for
Swiss-Prot version 24, where the sentence is replaced by ``the
active-site is not encoded by the opal codon uga but by ugc.''. This
replacement indicates that the knowledge in the original annotation is
now considered erroneous. Our definition of erroneous annotation
follows that of UniProt~\cite{UniProtNews}: An erroneous annotation is
one that is out of sync with respect to the biological knowledge.
Indeed, it may be that the original information is incorrect, rather
than the annotation.

Because of its nature, we can only detect transient sentences in the
release before the current; a sentence must be added and then removed
from the next release cycle. However, this pattern fits with previous
research that links annotation quality to
stability~\cite{Gross09Estimating}; annotations that are persistent
over many release cycles provide greater confidence and likelihood in
their correctness. Therefore, using this information we can conclude
that the introduction of an annotation within an entry update is more
likely to be volatile than those which have remained over numerous
releases. Importantly, transient sentences should not be seen as a
burden to the overall quality of a database but used to indicate the
importance of annotation maturity.

Although less common than transient sentences, over $8,500$ sentences
in Swiss-Prot appear to originate from TrEMBL. This is a surprising
observation; annotations in Swiss-Prot are considered manually
reviewed and curated. Further, TrEMBL annotations can be generated
based upon information from Swiss-Prot
annotations~\cite{Update13UniProt}. Although automated annotations on
the whole are typically of lesser quality than their manual
counterparts~\cite{Bell12Annotation}, as part of their incorporation
into Swiss-Prot, they will have undergone manual review. One possible
explanation for this is that, for a period of approximately two years,
some annotations in TrEMBL appear to have undergone manual
annotation~\cite{o2002high}. This was likely a result of a change in
annotation policy, and it is interesting that we are able to identify
such changes through this approach.

Clearly, a quality analysis between these sentences and those
originating directly from Swiss-Prot would be of value. However, this
analysis is not straightforward; no annotation quality metric that can
analyse individual annotations is available. This result does,
however, highlight that annotation provenance should be clearly
documented and available to users, especially given that research has
suggested that users often assume annotations are of a consistent
quality~\cite{Ussery04Genome}.

\begin{figure*}[!ht]
{\centering
 \includegraphics[width=\textwidth]{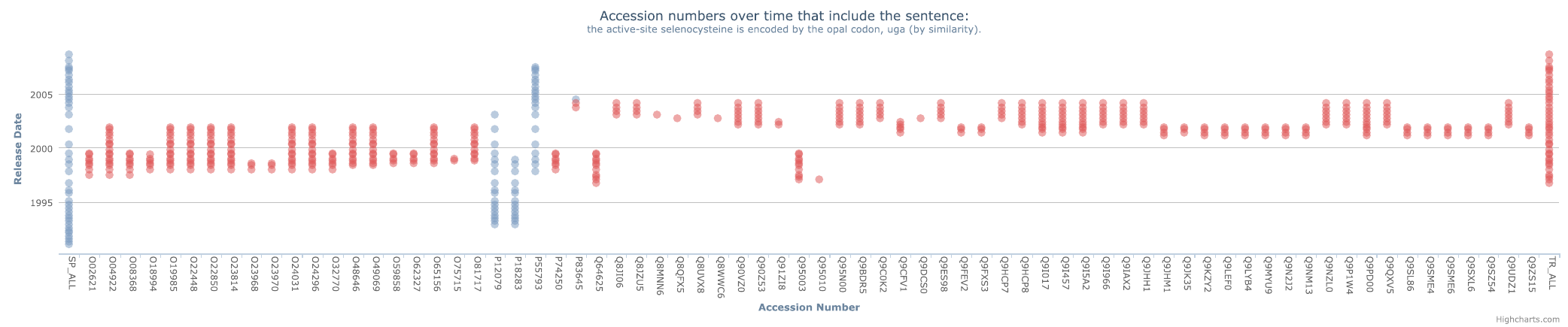}}
\caption{{\bf Visualising sentence propagation.} Visualisation for the
  sentence ``the active-site selenocysteine is encoded by the opal
  codon, uga (by similarity).''}
 \label{fig:selenoSimilarity}
 \end{figure*}

Another interesting pattern observed is from those sentences which are
removed from an entry, only to reappear in a subsequent release of the
same entry; i.e. they follow the reappearing sentences pattern. In
Figure~\ref{fig:interGraph} there are two examples of this pattern;
the sentence reappears after 7 years in entry P18283 and 11 years in
entry P12079. In these entries, the sentence was replaced with ``the
active-site selenocysteine is encoded by the opal codon, uga (by
similarity).'', with the visualisation for this sentences shown in
Figure~\ref{fig:selenoSimilarity}. The usage of ``by similarity''
suggests that the information is based on sequence similarity.
Interestingly, this sentence also follows the ``missing origin''
pattern.

Sentences exhibiting this pattern appear to indicate a conflict in the
underlying evidence and some uncertainty as to the correct
annotation. The impact of this pattern is similar to transient
sentences; they highlight the importance of annotation stability and
provenance.

The final pattern observed are sentences which are missing their root
origins.  Within Figure~\ref{fig:interGraph} the sentence initially
occurs within two entries and remains in these entries until
Swiss-Prot version 44, when it is subsequently removed. It is also
removed from the majority of the other entries, however still remains
in nine entries. Therefore, in Swiss-Prot Version 45 the sentence
exists in these nine entries when it has been removed from the entries
where it originated from. Depending on its reason for removal, it
could highlight that possibly erroneous annotation still remains in
the database.

However, as previously mentioned, this information was moved from the
textual annotation to the feature table in UniProtKB
entries. Therefore, this was not biologically erroneous in these nine
entries. However, it clearly should have been moved to the feature
table in all entries for consistency. This highlights how missing
propagation of textual annotation can lead to inconsistencies between
entries.

Changes in annotation are typically made to reflect an update in
knowledge; in light of new knowledge a previous annotation may now be
erroneous with respect to current knowledge. Given that annotations
propagate, any updates to an original annotation should also be
propagated. However, we identify over $8,000$ sentences which may, or
may have, incorrectly remained in the database.

While these first three patterns are of interest in regard to
annotation quality, we next investigate whether we can use the missing
root origin pattern as an indication for erroneous annotations.

\subsection*{Exploring the annotation space: Can we identify erroneous
  annotation?}

As discussed in the previous section, over 8000 sentences exhibiting
the missing origin pattern were identified. Here, we wish to analyse
this pattern to support our hypothesis that it can be used to identify
erroneous annotation. We define a missing origin sentence as one
which:

\begin{enumerate}
\item Initially occurs in the \emph{origin} entry.
\item Later appears in an additional entry; i.e. a \emph{secondary}
  entry.
\item Is removed or changed in the origin entry.
\item Remains unchanged within the secondary entry for a subsequent
  database release (or releases).
\end{enumerate}

Within this definition a sentence may also originate from, or
propagate, to multiple entries. We determined that each sentence can
be broadly categorised into one of five possible classifications:

\begin{itemize}
\item \textbf{Erroneous} The sentence in the secondary entry was
  inaccurate or incorrect given updates to the origin entry. This
  includes any case where detail is added or removed and not reflected
  in all relevant entries. This may include a sentence that has been
  reworded or one that has been removed entirely.

\item \textbf{Inconsistent} Whilst the sentence in the origin entry
  has been updated, it has not changed the biological information
  contained within the sentence, or been propagated to the secondary
  entries. The correction of a grammatical error would be an example.

\item \textbf{Accurate} The sentence in the secondary entry is
  accurate. Either the sentences appear identical by coincidence or
  the updates to the origin are not valid in the secondary entry.
  Therefore both annotations have become independent. For example,
  expression information may not be relevant in different organisms
  for the same gene.

\item \textbf{Too many results} The sentence was very heavily reused
  within UniProtKB and deemed infeasible to analyse. The more entries
  that a sentence occurs within, the more troublesome it becomes to
  classify individually, given the vast number of entries they occur
  in. Specifically, sentences that occurs in over 100 are classified
  as ``too many results''.

\item \textbf{Possibly erroneous} Some sentences did not carry enough
  evidence, or contained conflicting information, making a more
  confident decision of classification impossible.
  
\end{itemize}

There are four main decisions when evaluating the classification of a
textual annotation: deciding whether it is feasible to analyse the
sentence; determining whether the sentence appears to have been copied
between entries; deciding whether the update to the origin was
relevant to the secondary entry and deciding whether the update
affected the meaning of the textual annotation.

These decisions are subjective as interpretation of biological data
can vary between users. Whilst our protocol attempts to allow for
consistent interpretation, it is inevitable that reproducibility
cannot always be attained between different users. Given this, a
systematic and precise protocol was developed to encourage
reproducible results. This protocol, represented as a decision tree in
Figure~\ref{fig:decTree}, involves seven stages in determining a
sentences classification:

\begin{figure}[!ht]
\begin{center}
\includegraphics[width=0.45\textwidth]{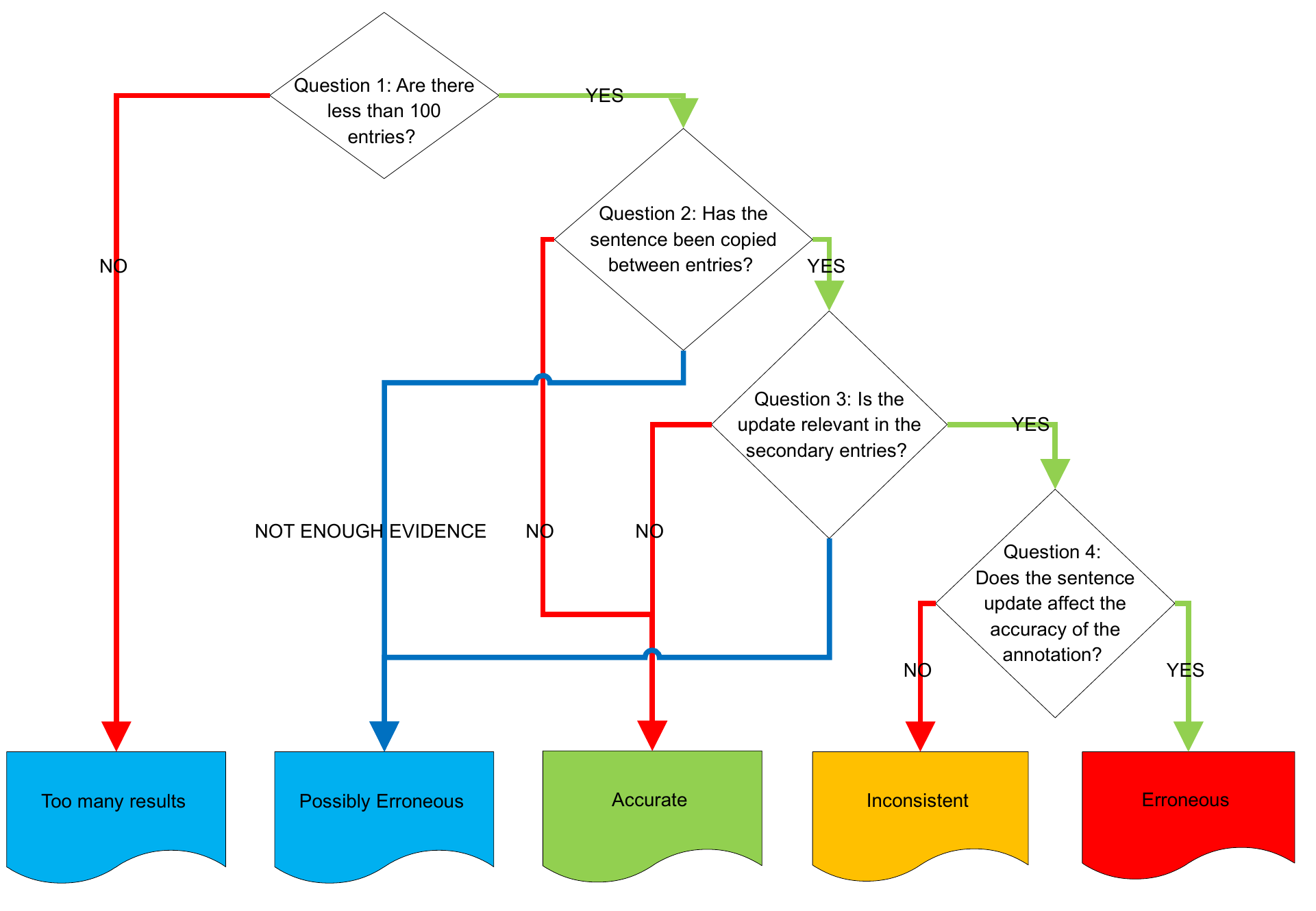}
\end{center}
\caption{ {\bf Decision tree summarising the protocol used to
    determine the classification of a sentence.} There are four main
  questions within the protocol that lead to a sentence being
  classified into one of five possible classifications.}
 \label{fig:decTree}
 \end{figure} 

\begin{enumerate}

\item Determine how many entries the sentence has propagated to. A
  sentence occurring in over 100 entries is infeasible to analyse
  (Figure~\ref{fig:decTree}, Question 1). 

\item Using the visualisation framework, identify both the
  \emph{origin} and \emph{secondary} entries that the sentence occurs
  in.

\item Using the UniSave tool, analyse the context of the sentence
  within the origin and secondary entries at the time that the
  sentence was initially added to the secondary entry. Does this
  context suggest the sentence was propagated between the entries
  (Figure~\ref{fig:decTree}, Question 2)? 

\item Determine the context for when the sentence was updated or deleted in
  the origin entry, then determine the context of the sentence in the
  secondary entry at the time when the sentence was deleted from the origin
  entry.

\item Is the update in the origin sentence relevant to the secondary
  entry (Figure~\ref{fig:decTree}, Question 3)?
  
\item Does the update in the origin entry affect the accuracy of the secondary
  entry (Figure~\ref{fig:decTree}, Question 4)?

\end{enumerate}

\begin{figure}[!ht]
 \begin{center}
 \includegraphics[width=0.45\textwidth]{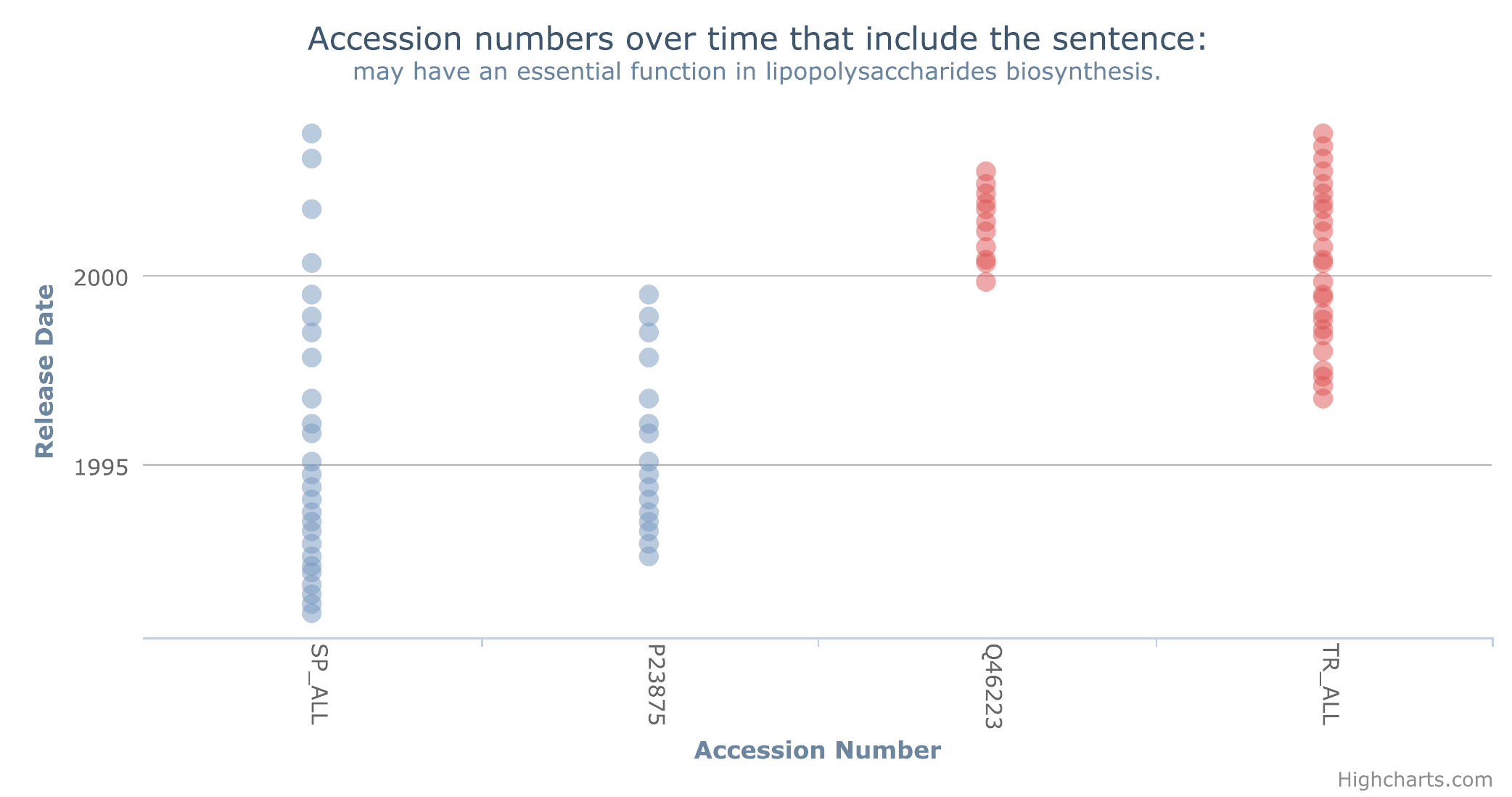}
 \end{center}
 \caption{{\bf Visualising sentence propagation.} Visualising the
   propagation of the sentence ``may have an essential function in
   lipopolysaccharides biosynthesis.'' through the database.}
 \label{fig:workedExample}
 \end{figure}

To illustrate this protocol, we can analyse the sentence ``may have an
essential function in lipopolysaccharides biosynthesis.'', for which
the associated visualisation is shown in
Figure~\ref{fig:workedExample}. This sentence appears in less than 100
entries (step 1), with the sentence originating in a single entry
(P23875) and being propagated to only a single secondary entry
(Q46223) (step 2). Analysing the context of the
origin\footnote{http://www.uniprot.org/uniprot/P23875.txt?version=11}
and secondary
entries\footnote{http://www.uniprot.org/uniprot/Q46223?version=7\&version=6}
at the time the sentence was added to the secondary entry shows
significant overlap (step 3). For example, information relating to
pathway information is also propagated. When the sentence is removed
from the origin
entry\footnote{http://www.uniprot.org/uniprot/P23875?version=11\&version=12},
the context appears relevant to the secondary entry (step 4). Given
this, it appears that the removal of this information should also be
applied to the secondary entry (step 5). Therefore, the sentence is
classified as erroneous (step 6). Indeed, this sentence is eventually
removed from the secondary
entry\footnote{http://www.uniprot.org/uniprot/Q46223?version=12\&version=11}.
When the sentence was removed from the origin entry (P23875) it was
replaced with a cautionary topic stating that it was initially
believed to have a function in lipopolysaccharides biosynthesis. This
suggests that an update in knowledge meant the old annotation is now
incorrect. When the sentence was removed from the secondary entry
(Q46223), after ten database releases (three years), it was removed
along with all other comments. Lipopolysaccharides biosynthesis was
also removed from the keyword list; the only reference to
lipopolysaccharides biosynthesis within the entry is in the title of a
referenced article. This suggests that the sentence could have been
removed ten database releases (three years) earlier.

Using this protocol, we analysed a total of $122$ sentences;
approximately $1.5\%$ of the $8355$ identified sentences. The complete
set of analysed sentences are shown at the end of the manuscript in
Table~\ref{tab:sentenceClass}, whilst the results are summarised in
Table~\ref{tab:missingOriginResults}. A number of these $122$
sentences were initially analysed to allow the protocol to be
developed and refined. Additionally, a number of sentences under 20
characters long were analysed. To remove any sentence length bias a
subset of sentences were taken; sentences were sorted by length and
every hundredth sentence over 20 characters long was analysed. The
decision to normalise for sentence length bias was based on the
assumption that longer sentences are more likely to have greater
information content. This resulted in $65$ sentences being analysed,
as summarised in Table~\ref{tab:missingOriginResultsLengthBias}.

\begin{table*}[!ht]
\centering
\caption{
\bf{Sentence classification results}}
{\small
\begin{tabular}{|l|l|l|l|l|l|}
  \hline
  \textbf{Classification} & \textbf{Erroneous} & \textbf{Inconsistent} & \textbf{Accurate} & \textbf{Too many Results} & \textbf{Possibly Erroneous} \\
  \hline
  Absolute & 36 & 29 & 28 & 15 & 14 \\
  \hline
  Percentage & 29.5\% & 23.8\% & 23.0\% & 12.3\% & 11.5\% \\
  \hline
  Potentially Erroneous & 2465 & 1986 & 1918 & 1027 & 959 \\
  \hline
\end{tabular}
}
\begin{flushleft}The classification results for all of the analysed
  sentences (122 in total).
\end{flushleft}
\label{tab:missingOriginResults}
\end{table*}

\begin{table*}[!ht]
\centering
\caption{
\bf{Classification of sentences over 20 characters in length}}
{\small
\begin{tabular}{|l|l|l|l|l|l|}
  \hline
  \textbf{Classification} & \textbf{Erroneous} & \textbf{Inconsistent} & \textbf{Accurate} & \textbf{Too many Results} & \textbf{Possibly Erroneous} \\
  \hline
  Absolute & 16 & 11 & 20 & 5 & 13 \\
  \hline
  Percentage & 24.6\% & 16.9\% & 30.8\% & 7.7\% & 20.0\% \\
  \hline
  Potentially Erroneous & 2057 & 1414 & 2571 & 643 & 1671 \\
  \hline
\end{tabular}
}
\begin{flushleft}The classification results of the 65 sentences analysed,
  controlling for sentence length bias (i.e. every 100th sentence over
  20 characters in length). 
\end{flushleft}
\label{tab:missingOriginResultsLengthBias}
\end{table*}

Our results show that $42\%$ of sentences were identified as erroneous
or inconsistent. Thirteen sentences were classified as ``possibly
erroneous''; in these instances we believed there was not enough
evidence to confidently make a reasoned decision. This mostly arose
when trying to determine sentence context (Figure~\ref{fig:decTree},
question 2). A similar number of sentences were classified as
``inconsistent'', which suggests the curation process is
asynchronized. The changes to these sentences were typically
grammatical, often corrected after a number of versions. These issues
could be overcome, or substantially reduced, if formal provenance were
available.

Of these $65$ analysed sentences, 32 sentences remain in the most
recent UniProtKB database release, with this subset of results
summarised in Table~\ref{tab:missingOriginResultsRecentDB}. A
breakdown of these results is provided on the supporting website.

\begin{table*}[!ht]
\centering
\caption{
\bf{Classification of sentences in UniProtKB 2012\_05}}
{\small
\begin{tabular}{|l|l|l|l|l|l|}

  \hline
  \textbf{Classification} & \textbf{Erroneous} & \textbf{Inconsistent} & \textbf{Accurate} & \textbf{Too many Results} & \textbf{Possibly Erroneous} \\
  \hline
  Absolute & 4 & 5 & 12 & 1 & 10 \\
  \hline
  Percentage & 12.5\% & 15.6\% & 37.5\% & 3.1\% & 31.3\% \\
  \hline
  Potentially Erroneous & 479 & 599 & 1438 & 120 & 1198 \\
  \hline

\end{tabular}
}
\begin{flushleft}The classification of results for the subset of
  sentences controlling for sentence length bias analysed that remain
  in UniProtKB Version 2012\_05. 
\end{flushleft}
\label{tab:missingOriginResultsRecentDB}
\end{table*}

Following up from these findings, we contacted the UniProtKB helpdesk
to query our results. We submitted a detailed breakdown of three
sentences to see if they agreed with our classification. For two of
the three sentences, which are historical, they confirmed that if the
sentence was to be re-added to the entry it would now be considered
incorrect. For the final sentence, in the extant database, the origin
sentence has been modified, and the current biological knowledge is
not rich enough to determine whether the secondary entry is accurate;
however, we suggest, our analysis raises a sensible question, that
should be addressed as knowledge increases.

\section*{Discussion}

Current methods for detecting textual annotation provenance and
tracking its propagation are somewhat limited. Within this paper we
have presented a technique that allows annotation provenance and
propagation to be identified and visualised by using
sentences. Further, we have provided an analysis of sentence reuse
levels and identified a number of annotation patterns that provide an
indication as to an annotations quality and correctness.

The cornerstone of this work was dependent upon sentence reuse in
UniProtKB. Our analysis shows that reuse is heavily prevalent for both
Swiss-Prot and TrEMBL. This is because of the curation process
employed by UniProt~\cite{Magrane11UniProt}, which consists of six key
stages, one of which involves identifying similar entries and
standardising annotations between these entries. If two entries from
the same gene and species are identified then they are merged.
Therefore, sentences are effectively copied between entries as a
matter of protocol. This process can see sections, or sometimes whole
annotations, from one entry being copied to other entries without
change.

Our analysis shows that this curation approach has decreased both the
percentage of entries without annotation and increased the average
number of sentences per entry over time. Whilst this appears to
indicate an improvement in annotation coverage, in line with their
stated goals, the actual annotation corpus is becoming more
duplicated. We have shown that the average number of entries each
sentence appears in is increasing, with the percentage of unique
sentences in the latest version of UniProtKB being $< 7\%$ for
Swiss-Prot and $<0.03\%$ for TrEMBL. Similar patterns of reuse are
also shown in work by Schnoes \emph{et al.}, who have shown that the
annotation of many high-throughput experiments is based upon a very
small amount of experimental data~\cite{Schnoes13Biases}.

Whilst the levels of reuse are generally increasing overtime, we
interestingly note a slight decline in sentence reuse for later
versions of Swiss-Prot. Although this decline coincides with the
change of the UniProtKB release cycle, it appears to be related to a
change in annotation policy for Swiss-Prot. After 2010 only sequences
with experimental annotation were added to Swiss-Prot; previously
automatically annotated orthologue sequences from complete genomes
were often included in Swiss-Prot.

In the face of ever increasing raw biological data, this reuse is not
unexpected. Whilst manual curation is often regarded as the `gold
standard'~\cite{Curwen04Ensembl}, it is a significant bottleneck. For
example, in the FlyBase database it can take between two and four
months for an article to be manually curated with consideration
recently being given to incorporating sections of automated processing
into the curation process~\cite{McQuilton12FlyBase}. It was for this
same reason that UniProtKB introduced TrEMBL in 1996. Whilst reuse is
understandably higher within automated methods, it is inevitably going
to remain commonplace throughout both automated and manual databases
while the quantity of raw biological data being generated continues to
increase. Indeed, sentence reuse is an important feature of annotation
curation. In addition to the propagation of knowledge, it also allows
annotations to become standardised and can be used to enforce levels
of quality control.

Whilst these results further highlight the importance of being able to
identify the origin of an annotation, the analysis was only achievable
given that UniProtKB make available all major historical versions of
Swiss-Prot and TrEMBL. Users are typically only interested in the most
recent and up-to-date biological data available, but this work
highlights the added value and importance of being able to scour
archival data; database features such as UniSave should be a
requirement rather than a luxury.

It was this archival data that allows provenance and propagation to be
analysed, allowing the development of a visualisation technique. These
visualisations appear to be useful, as their usage allowed a number of
propagation patterns to be identified.

Provenance is inferred by identifying the first UniProtKB entry that a
sentence appears in. Similarly, the propagation of a sentence is
viewed by determining all subsequent UniProtKB entries the sentence
appears in over time. For individual sentences, this inference is not
necessarily correct. For example, a sentence may originate in an entry
outside of UniProtKB or within a minor release. Further, the appearance
of a sentence in multiple entries may be an independent event, with no
relationship between the entries. However, the curation process and
levels of reuse identified would argue against this often being the
case.  More formal tracking of provenance within the database curation
process would help to alleviate this difficulty.

Confidence that the apparent propagation of a sentence is correct can
be gained by analysing the context that the sentences appear in, for
example, by comparing sequence similarity. Indeed, performing such an
analysis on the sentence ``the active-site selenocysteine is encoded
by the opal codon, uga.'' led to the identification of four
propagation patterns.

We believe that these identified patterns hold promise as quality and
correctness indicators. For example, a sentence which adheres to the
``reappearing entry'' pattern could be considered more dubious, as its
inclusion (or exclusion) within an entry is not definitive. Further,
although not shown by the visualisations, a number of entries sharing
a sentence have later become merged. For example, in the latest
version of UniProtKB, accessions P22352; O43787; Q86W78; Q9NZ74 and
Q9UEL1 are all merged into a single entry, with a sentence common to
all entries remaining in the merged entry. Our analysis identified a
number of sentences that adhere to each of the four patterns. These
patterns were identified by manual inspection during the development
of the visualisation framework. Further work could be undertaken to
perform a comprehensive search to identify any additional propagation
patterns. This work could also be extended to derive a quantitative
metric. By combining these results with other textual metrics, such as
Inverse Document Frequency (IDF)~\cite{Jones72IDF}, annotations
between entries could be scored and rated.

Deriving a quality metric is not straightforward. However, we
hypothesised that the ``missing origin'' pattern could be used to
identify erroneous annotations. This analysis is more discrete than
deriving a quality measure, as a sentence can be classified into one
of five groups. Our analysis identified a number of annotations we
believe to be erroneous, including a number of sentences that still
remain in the latest version of UniProtKB. As acknowledged earlier,
these results are somewhat subjective. However, the UniProt help desk
have checked our conclusions for three cases; in two cases these were
correct, and in the third, we lack the biological knowledge to draw a
definitive conclusion.

These results suggests that the identification of propagation patterns
could aid in the discovery of erroneous annotations, and act as a
mechanism to increase confidence into an annotations quality.

Within this paper we only analysed sentence propagation between major
UniProtKB versions. UniProtKB versions prior to Version 2010\_01 made
the distinction between minor and major releases. Minor versions are
not archived on the UniProtKB FTP server, but can be viewed
interactively via the UniSave tool. A finer level of granularity could
be achieved by extracting this information from UniSave and
incorporating into our tool. This could unearth additional sentences
that fit propagation patterns and may help distinguish the provenance
of a sentence that appears to originate in two or more entries. In
practice, however, this would be complex as the version numbers of
UniSave and UniProtKB differ; exacerbating the problems caused by the
lack of coordination between Swiss-Prot and TrEMBL releases.

This work could also be extended to analyse the evolution of
individual sentences. Whilst sentences can be copied verbatim between
entries, many sentences will be copied and then undergo minimalistic
changes, such as the change of a single letter or word. By employing
semantic similarity~\cite{Lord03Semantic} coupled with a combination
of historical data and IDF it may be possible to identify and track
sentence, and annotation, evolution.

The structure and features of UniProtKB made it an ideal resource to
perform this analysis. A clear extension to this work would be to
apply the techniques and tools within this paper to other databases,
allowing propagation and provenance to be identified. As noted
earlier, it is plausible that annotations propagate \emph{between}
databases. For example, the InterPro database is used in the
production of TrEMBL~\cite{Apweiler01InterPro}, whilst the neXtProt
database integrates the annotation in UniProtKB/Swiss-Prot as a
primary source, as well as incorporating data from a number of other
sources such as GO and Ensembl~\cite{Lane12neXtProt}. With over 1500
active biological databases~\cite{Fernandez13Database}, if
cross-database propagation does indeed occur, then the provenance map
could be vast, and using this approach it is plausible that the
``true'' provenance and propagation of an annotation could be
identified, and used to increase the quality of all databases. The
visualisation tool was developed in a manner that will allow any
textual resource to be compared.

Our initial analysis has provided a number of fruitful results.
Extending this analysis to cross-database propagation and provenance
could provide even more encouraging results and could take a
significant step towards the ability to track and trace annotation
propagation.

\section*{Acknowledgments}

We thank the UniProt helpdesk for their help in answering our queries.
MJB is funded by EPSRC. MC is funded by EPSRC grant (EP/G037620/1) for
the Engineering Doctorate in Biopharmaceutical Process Development.
The funders had no role in study design, data collection and analysis,
decision to publish, or preparation of the manuscript.

\bibliographystyle{natbib}
\bibliography{Sentence_Reuse_Bell_Lord}

\onecolumn

\begin{longtable}[h]{|p{12cm}|l|}
\caption{\textbf{All analysed sentences and their classification}} \\
\hline
\textbf{Sentence} & \textbf{Classification} \\
\hline
\endfirsthead
\hline
\textbf{Sentence} & \textbf{Classification} \\
\endhead
\hline
\multicolumn{2}{r}{\textit{Continued on next page}} \\
\endfoot
\hline
\endlastfoot
belongs to the 40s cdc5-associated complex (or cwf complex), a spliceosome sub-complex reminiscent of a late-stage spliceosome composed of the u2, u5 and u6 snrnas and at least brr2, cdc5, cwf2, cwf3, cwf4, cwf5, cwf6, cwf7, cwf8, cwf9, cwf10, cwf11, cwf12, cwf13, cwf14, cwf15, cwf16, cwf17, cwf18, cwf19, cwf20, cwf21, cwf22, cwf23, cwf24, cwf25, cwf26, cwf27, cwf28, ist3, lea1, msl1, prp5, prp10, prp12, prp17, prp22, sap61, sap62, sap114, sap145, slu7, smb1, smd1, smd3, smf1, smg1 and syf2. & Inconsistent \\ \hline
the light chain is composed of three structural domains: a large globular n-terminal domain which may be involved in binding to kinesin heavy chains, a central alpha-helical coiled-coil domain that mediates the light chain dimerization; and a small globular c-terminal which may play a role in regulating mechanochemical activity or attachment of kinesin to membrane-bound organelles (by similarity). & Erroneous \\ \hline
the biological conversion of cellulose to glucose generally requires three types of hydrolytic enzymes: 1) endoglucanases which cut internal beta-1,4-glucosidic bonds; 2) exocellobiohydrolases that cut the dissaccharide cellobiose from the nonreducing end of the cellulose polymer chain; 3) beta-1,4-glucosidases which hydrolyze the cellobiose and other short cello-oligosaccharides to glucose. & Inconsistent \\ \hline
in the hair cortex, hair keratin intermediate filaments are embedded in an interfilamentous matrix, consisting of hair keratin-associated protein (krtap), which are essential for the formation of a rigid and resistant hair shaft through their extensive disulfide bond cross-linking with abundant cysteine residues of hair keratins. & Inconsistent \\ \hline
the beta subunit of voltage-dependent calcium channels contributes to the function of the calcium channel by increasing peak calcium current, shifting the voltage dependencies of activation and inactivation, modulating g protein inhibition and controlling the alpha-1 subunit membrane targeting (by similarity). & Erroneous \\ \hline
interacts with the c-terminal of peptidylglycine alpha-amidating monooxygenase (pam) and may act as part of a signal transduction system linking the catalytic domains of pam in the lumen of the secretory pathway to cytosolic factors regulating the cytoskeleton and signal transduction pathways. & Erroneous \\ \hline
the modification is dependent on dna and is involved in the regulation of various important cellular processes such as differentiation, proliferation, and tumor transformation and also in the regulation of the molecular events involved in the recovery of cell from dna damage (by similarity). & Erroneous \\ \hline
adenosylhomocysteine is a competitive inhibitor of s-adenosyl-l-methinine-dependent methyl transferase reactions; therefore adenosylhomocysteinase may play a key role in the control of methylations via regulation of the intracellular concentration of adenosylhomocysteine (by similarity). & Inconsistent \\ \hline
component of the multisynthetase complex which is comprised of a bifunctional glutamyl-prolyl-trna synthetase, the monospecific isoleucyl, leucyl, glutaminyl, methionyl, lysyl, arginyl, and aspartyl-trna synthetases as well as three auxiliary proteins, p18, p48 and p43 (by similarity). & Erroneous \\ \hline
self; 2; ebi-311928, ebi-311928; p03949:abl-1; 4; ebi-311928, ebi-2315883; q17539:c01b10.8; 5; ebi-311928, ebi-311920; q95qi7:daf-3; 2; ebi-311928, ebi-326363; q09248:dnc-2; 2; ebi-311928, ebi-316282; q09975:lys-8; 2; ebi-311928, ebi-313861; q21831:snfc-5; 2; ebi-311928, ebi-360213; & Erroneous \\ \hline
the n-terminal of the protein extends into the stroma where it is involved with adhesion of granal membranes and photoregulated by reversible phosphorylation of its threonine residues; both are believed to mediate the distribution of excitation energy between photosystems i and ii. & Inconsistent \\ \hline
the modification is dependent on dna and is involved in the regulation of various important cellular processes such as differentiation, proliferation, and tumor transformation and also in the regulation of the molecular events involved in the recovery of cell from dna damage. & Erroneous \\ \hline
the iicd domains contain the sugar binding site and the transmembrane channel; the iia domain contains the primary phosphorylation site (the donor is phospho-hpr); iia transfers its phosphoryl group to the iib domain which finally transfers it to the sugar (by similarity). & Too Many Results \\ \hline
adenosylhomocysteine is a competitive inhibitor of s-adenosyl-l-methinine-dependent methyl transferase reactions; therefore adenosylhomocysteinase may play a key role in the control of methylations via regulation of the intracellular concentration of adenosylhomocysteine. & Inconsistent \\ \hline
this delta-9 desaturase is a terminal component of the liver microsomal stearyl-coa desaturase system, that utilizes o(2) and electrons from reduced cytochrome b(5) to catalyze the insertion of a double bond into a spectrum of fatty acyl-coa substrates (by similarity). & Inconsistent \\ \hline
in the absence of mercury merr represses transcription by binding tightly to the mer operator region; when mercury is present the dimeric complex binds a single ion and becomes a potent transcriptional activator, while remaining bound to the mer site (by similarity). & Erroneous \\ \hline
chemotactic-signal tranducers respond to changes in the concentration of attractants and repellents in the environment, transduce a signal from the outside to the inside of the cell, and facilitate sensory adaptation through the variation of the level of methylation. & Inconsistent \\ \hline
activated by tyrosine-phosphorylation in response to either integrin clustering induced by cell adhesion or antibody cross-linking, or via g-protein coupled receptor (gpcr) occupancy by ligands such as bombesin or lysophosphatidic acid, or via ldl receptor occupancy. & Erroneous \\ \hline
laminin is a complex glycoprotein, consisting of three different polypeptide chains (alpha, beta, gamma), which are bound to each other by disulfide bonds into a cross-shaped molecule comprising one long and three short arms with globules at each end (by similarity). & Erroneous \\ \hline
psi is a plastocyanin-ferredoxin oxidoreductase, converting photonic excitation into a charge separation, which transfers an electron from the donor p700 chlorophyll pair to the spectroscopically characterized acceptors a0, a1, fx, fa and fb in turn (by similarity). & Erroneous \\ \hline
involved in protection of chromosomal dna from damage under nutrient-limited and oxidative stress conditions. & Inconsistent \\ \hline
belongs to the cold-shock domain (csd) family. & Too Many Results \\ \hline
p35415:prm; 1; ebi-86215, ebi-133215; & Erroneous \\ \hline
composed of 14 different subunits. & Possibly Erroneous \\ \hline
proteins that associate with the core dimer include three families of regulatory subunits b (the r2/b/pr55/b55, r3/b''/pr72/pr130/pr59 and r5/b'/b56 families), the 48 kda variable regulatory subunit, viral proteins, and cell signaling molecules (by similarity). & Inconsistent \\ \hline
type i restriction and modification enzymes are complex, multifunctional systems which require atp, s-adenosyl methionine and mg(2+) as cofactors and, in addition to their endonucleolytic and methylase activities, are potent dna-dependent atpases (by similarity). & Inconsistent \\ \hline
3-beta-hydroxy-delta(5)-steroid + nad(+) = 3-oxo-delta(5)-steroid + nadh (acts on 3-beta-hydroxyandrost-5-en-17-one to form androst-4-ene-3,17-dione and on 3-beta-hydroxypregn -5-en-20-one to form progesterone). & Accurate \\ \hline
udp-n-acetyl-d-glucosamine + n-acetyl-beta-d-glucosaminyl-1,2-alpha-d-mannosyl-1,3(6)-(n-acetyl-beta-d-glucosaminyl-1,2-alpha-d-mannosyl,1,6(3))-beta-d-mannosyl-1,4-n-acetyl-beta-d-glucosaminyl-r = udp + n-acetyl-beta-d-glucosaminyl-1,2-(n-acetyl-beta-d-glucosaminyl-1,6)-1,2-alpha-d-mannosyl-1,3(6) -(n-acetyl-beta-d-glucosaminyl-1,2-alpha-d-mannosyl-1,6(3))-beta-d-mannosyl-1,4-n-acetyl-beta-d-glucosaminyl-r. & Erroneous \\ \hline
in e.coli rnase h participare in dna replication; it helps to specify the origin of genomic replication by suppressing initiation at origins other than the locus oric; along with the 5\'-3\' exonuclease of pol1, it removes rna primers from the okazaki fragments of lagging strand symthesis; and it defines the origin of replication for cole1-type plasmids by specific cleavage of an rna preprimer. & Inconsistent \\ \hline
thoracic aortic aneurysms and dissections are primarily associated with a characteristic histologic appearance known as \'medial necrosis\' or \'erdheim cystic medial necrosis\' in which there is degeneration and fragmentation of elastic fibers, loss of smooth muscle cells, and an accumulation of basophilic ground substance. & Erroneous \\ \hline
component of the cleavage and polyadenylation specificity factor (cpsf) complex that play a key role in pre-mrna 3\'-end formation, recognizing the aauaaa signal sequence and interacting with poly(a) polymerase and other factors to bring about cleavage and poly(a) addition (by similarity). & Inconsistent \\ \hline
there are two operons: the xylcab operon is responsible for the upper metabolic pathway from toluene to aromatic carboxylic acids, \& the xyldlefg operon is required for the lower catabolic pathway from aromatic carboxylic acids to compounds that enter the trycarboxylic acid cycle. & Erroneous \\ \hline
hh is characterized by abnormal intestinal iron absorption and progressive increase of total body iron, which results in midlife in clinical complications including cirrhosis, cardiopathy, diabetes, endocrine dysfunctions, arthropathy, and susceptibility to liver cancer. & Inconsistent \\ \hline
prp is found in high quantity in the brain of humans and animals infected with the degenerative neurological diseases kuru, creutzfeldt-jacob disease (cjd), gerstmann-straussler syndrome (gss), scrapie, bovine spongiform encephalopathy (bse), etc. to other prp. & Accurate \\ \hline
involved in the atp-dependent selective degradation of cellular proteins, the maintenance of chromatin structure, the regulation of gene expression, the stress response, and ribosome biogenesis (by similarity). & Erroneous \\ \hline
coup (chicken ovalbumin upstream promoter) transcription factor binds to the ovalbumin promoter and, in cunjunction with another protein (s300-ii) stimulates initiation of transcription. & Inconsistent \\ \hline
the lys-124 ubiquitination also modulates the formation of double-strand breaks during meiosis and is a prerequisite for and dna-damage checkpoint activation (by similarity). & Erroneous \\ \hline
the export to cytoplasm depends on the interaction with a 14-3-3 chaperone protein and is due to its phosphorylation at ser-259 and ser-498 by camk (by similarity). & Erroneous \\ \hline
the sigma factor is an initiation factor that promotes attachment of the rna polymerase to specific initiation sites and then is released (by similarity). & Too Many Results \\ \hline
hydrolysis of 1,4-alpha-d-glucosidic linkages in polysaccharides so as to remove successive maltose units from the non-reducing ends of the chains. & Accurate \\ \hline
the resulting products may subsequently be converted to the corresponding alcohols that are incorporated into lignins (by similarity). & Erroneous \\ \hline
involved in the initial immune cell clustering during inflammatory response and may regulate chemotactic activity of chemokines. & Inconsistent \\ \hline
s-adenosyl-l-methionine + magnesium protoporphyrin = s-adenosyl-l-homocysteine + magnesium protoporphyrin monomethyl ester. & Erroneous \\ \hline
component of the coat surrounding the cytoplasmic face of coated vesicles located at the golgi complex (by similarity). & Accurate \\ \hline
hsp82 is an essential protein that is required by cells in higher concentrations for growth at higher temperatures. & Accurate \\ \hline
monoubiquitinated on lys-147; may give a specific tag for epigenetic transcriptional activation (by similarity). & Erroneous \\ \hline
probably a dodecamer composed of six biotin-containing alpha subunits and six beta subunits (by similarity). & Possibly Erroneous \\ \hline
organized into a structure (processome or rna degradosome) containing a number of rna-processing enzymes. & Inconsistent \\ \hline
involved in the formation of the nuclear envelope and of the transitional endoplasmic reticulum (ter). & Inconsistent \\ \hline
this methionine-rich region is probably important for copper tolerance in bacteria (by similarity). & Erroneous \\ \hline
they have identical ligand binding properties but different coupling properties with g proteins. & Possibly Erroneous \\ \hline
3-carboxy-2-hydroxy-4-methylpentanoate + nad(+) = 3-carboxy-4-methyl-2- oxopentanoate + nadh. & Accurate \\ \hline
this is a conceptual translation; two frameshifts had to be introduced to produce this orf. & Erroneous \\ \hline
component of the infraciliary lattice (icl) and the ciliary basal bodies (by similarity). & Possibly Erroneous \\ \hline
catalyzes the methylation of c-11 in precorrin-4 to form precorrin-5 (by similarity). & Possibly Erroneous \\ \hline
on the 2d-gel the determined pi of this unknown protein is: 6.2, its mw is: 28 kda. & Accurate \\ \hline
heterodimer of a p110 (catalytic) and a p85 (regulatory) subunit (by similarity). & Accurate \\ \hline
this viral protein may be involved in the regulation of the complement cascade. & Inconsistent \\ \hline
two forms; long (shown here) and short; are produced by alternative splicing. & Inconsistent \\ \hline
assembles at the inner surface of the cytoplasmic membrane (by similarity). & Too Many Results \\ \hline
1-aminocyclopropane-1-carboxylate + o2 = ethylene + hcn + co(2) + 2 h(2)o. & Accurate \\ \hline
bind preferentially single-stranded dna and unwind double stranded dna. & Inconsistent \\ \hline
involved in the regulation of hydrogenase expression (by similarity). & Erroneous \\ \hline
may have an essential function in lipopolysaccharides biosynthesis. & Erroneous \\ \hline
rch(2)nh(2) + h(2)o + acceptor = rcho + nh(3) + reduced acceptor. & Accurate \\ \hline
subunit 1 binds to the primer-template junction (by similarity). & Inconsistent \\ \hline
to immunoglobulin and major histocompatibility complex domain. & Too Many Results \\ \hline
isoform 3: membrane; multi-pass membrane protein (potential). & Possibly Erroneous \\ \hline
the beta subunit seems to be encoded by a multigene family. & Erroneous \\ \hline
atp + adenylylsulfate = adp + 3\'-phosphoadenylylsulfate. & Inconsistent \\ \hline
an aryl sulfate + a phenol = a phenol + an aryl sulfate. & Erroneous \\ \hline
peptidyl-l-amino acid + h(2)o = peptide + l-amino acid. & Possibly Erroneous \\ \hline
in the c-terminus to yeast sla2 and c.elegans zk370.3. & Erroneous \\ \hline
mediates e2-dependent ubiquitination (by similarity). & Accurate \\ \hline
villin is a ca(2+)-regulated actin-binding protein. & Inconsistent \\ \hline
atp + undecaprenol = adp + undecaprenyl phosphate. & Accurate \\ \hline
aminoacyl-peptide + h(2)o = amino acid + peptide. & Inconsistent \\ \hline
to the calcitonin and to the secretin receptors. & Erroneous \\ \hline
heterodimer of an alpha chain and a beta chain. & Too Many Results \\ \hline
requires ca2+ and mn2+ ions for full activity. & Inconsistent \\ \hline
contains 1 immunoglobulin-like v-type domain. & Too Many Results \\ \hline
belongs to family 13 of glycosyl hydrolases. & Too Many Results \\ \hline
acts as a transglycosylase (by similarity). & Erroneous \\ \hline
nuclear effector molecule (by similarity). & Possibly Erroneous \\ \hline
involved in carbon catabolite repression. & Erroneous \\ \hline
q9vy42:cg1461; 1; ebi-194476, ebi-127720; & Erroneous \\ \hline
contains 6 ldl-receptor class b domains. & Erroneous \\ \hline
ring cleavage of 2,3-dihydroxybiphenyl. & Possibly Erroneous \\ \hline
not expected to have protease activity. & Accurate \\ \hline
secreted in hemolymph (by similarity). & Accurate \\ \hline
interacts with rad51 (by similarity). & Accurate \\ \hline
endplasmic reticulum membrane bound. & Accurate \\ \hline
associated with the plasma membrane. & Accurate \\ \hline
does not have a catalytic activity. & Possibly Erroneous \\ \hline
belongs to the eae/invasin family. & Erroneous \\ \hline
interacts with cyclin g in vitro. & Possibly Erroneous \\ \hline
self; 1; ebi-190958, ebi-190958; & Possibly Erroneous \\ \hline
binds 1 nickel ion per monomer. & Accurate \\ \hline
binds 1 magnesium per subunit. & Inconsistent \\ \hline
clavulanic acid biosynthesis. & Accurate \\ \hline
belongs to the ycf50 family. & Accurate \\ \hline
inhibited by acetazolamide. & Erroneous \\ \hline
involved in tumorigenesis. & Accurate \\ \hline
acetyltransferase enzyme. & Possibly Erroneous \\ \hline
phosphorylates ppp1r12a. & Possibly Erroneous \\ \hline
detected at low levels. & Accurate \\ \hline
interacts with trim28. & Accurate \\ \hline
contacts protein l19. & Erroneous \\ \hline
interacts with gcn5. & Accurate \\ \hline
may self-associate. & Accurate \\ \hline
secreted in milk. & Too Many Results \\ \hline
heme-thiolate. & Accurate \\ \hline
adipocytes. & Accurate \\ \hline
nadp. & Accurate \\ \hline
nuclear. & Too Many Results \\ \hline
p. & Too Many Results \\ \hline
25. & Too Many Results \\ \hline
1. & Too Many Results \\ \hline
3. & Too Many Results \\ \hline
2. & Too Many Results \\ \hline
venom. & Inconsistent \\ \hline
roots. & Inconsistent \\ \hline
leaf. & Inconsistent 

\label{tab:sentenceClass}
\end{longtable}
\begin{flushleft} All of sentences analysed, and their corresponding
  classification. Sentences have been stored in lowercase to allow for
  case insensitive comparison. For further information, including the
  entries affected by these sentences, please see the authors website.
\end{flushleft}

\end{document}